\documentclass[letterpaper]{article} 
\usepackage{aaai25}  
\usepackage{times}  
\usepackage{helvet}  
\usepackage{courier}  
\usepackage[hyphens]{url}  
\usepackage{graphicx} 
\usepackage{natbib}  
\usepackage{caption} 
\usepackage{algorithm}
\usepackage{algorithmic}

\usepackage{amsfonts}
\usepackage{booktabs} 
\usepackage{subcaption}
\usepackage{colortbl}
\usepackage{amsmath}
\usepackage{subfloat}
\usepackage{enumitem}

\usepackage{newfloat}
\usepackage{listings}
\DeclareCaptionStyle{ruled}{labelfont=normalfont,labelsep=colon,strut=off} 
\floatstyle{ruled}
\newfloat{listing}{tb}{lst}{}
\floatname{listing}{Listing}
\title{Bidirectional Logits Tree: \\
Pursuing Granularity Reconcilement in Fine-Grained Classification}
\author {
    Zhiguang Lu\textsuperscript{\rm 1,\rm 2},
    Qianqian Xu\textsuperscript{\rm 1\thanks{Corresponding Authors}}, 
    Shilong Bao\textsuperscript{\rm 2}, 
    Zhiyong Yang\textsuperscript{\rm 2}, 
    Qingming Huang\textsuperscript{\rm 1,\rm 2,\rm 3*}
}
\affiliations {
    \textsuperscript{\rm 1}Key Laboratory of Intelligent Information Processing, Institute of Computing Technology, Chinese Academy of Sciences \\
    \textsuperscript{\rm 2}School of Computer Science and Technology, University of Chinese Academy of Sciences \\
    \textsuperscript{\rm 3}Key Laboratory of Big Data Mining and Knowledge Management, University of Chinese Academy of Sciences \\
    \{luzhiguang23s, xuqianqian\}@ict.ac.cn, \{baoshilong, yangzhiyong21, qmhuang\}@ucas.ac.cn
}

\begin{document}

\maketitle

\begin{abstract}
This paper addresses the challenge of \textit{Granularity Competition} in fine-grained classification tasks, which arises due to the semantic gap between multi-granularity labels. Existing approaches typically develop independent hierarchy-aware models based on shared features extracted from a common base encoder. 
However, because coarse-grained levels are inherently easier to learn than finer ones, the base encoder tends to prioritize coarse feature abstractions, which impedes the learning of fine-grained features.
To overcome this challenge, we propose a novel framework called the Bidirectional Logits Tree (BiLT) for \textit{Granularity Reconcilement}. 
The key idea is to develop classifiers sequentially from the finest to the coarsest granularities, rather than parallelly constructing a set of classifiers based on the same input features. 
In this setup, the outputs of finer-grained classifiers serve as inputs for coarser-grained ones, facilitating the flow of hierarchical semantic information across different granularities. 
On top of this, we further introduce an Adaptive Intra-Granularity Difference Learning (AIGDL) approach to uncover subtle semantic differences between classes within the same granularity. 
Extensive experiments demonstrate the effectiveness of our proposed method.
\end{abstract}

%
\begin{links}
    \link{Code}{https://github.com/ZhiguangLuu/BiLT}
\end{links}


\section{Introduction}
Fine-grained visual classification (FGVC) has been a longstanding and challenging research focus in the deep learning community \cite{HyperbolicLearningfromcoarse, Multimodal_Prompting_FGVC, tip/PuHWFDH24}. 
Unlike traditional image classification tasks, FGVC demands models to capture subtle distinctions between categories for accurate predictions. 
Recently, several studies \cite{zhang2024hls, chen2022label, wang2021hierarchical} have recognized that the hierarchical label structure (HLS) inherent in class names \cite{deng2009imagenet, krizhevsky2009learning, zhao2011large, maji2013fine} can significantly enhance the understanding and performance of FGVC tasks. 
Consequently, substantial efforts have been made to explore methods for effectively integrating hierarchical information into FGVC tasks \cite{jain2024test, mm/WangZZZJ23}. 
Generally, once hierarchical semantic information is incorporated, the primary goal is to ensure that the model’s predictions consistently align with the inherent hierarchical structure of the data. 
In other words, these methods aim to make precise predictions or at least keep the results as close as possible to semantically similar categories even when errors occur, thereby minimizing the severity of mistake \cite{bertinetto2020making}. 
For example, suppose a model misclassifies a ``Husky''. 
In that case, it is more acceptable within the hierarchy for the output to be another dog breed, such as a ``Corgi'', rather than an unrelated category like a ``Siamese Cat''.

In pursuit of this goal, one effective paradigm \cite{silla2011survey, zhang2013review} usually adopts a common shared encoder to extract the semantic features of images, and then \textbf{constructs hierarchy-aware models independently} for the classification tasks at different levels. 
Typically, \cite{bertinetto2020making} introduces a multi-task learning method that employs hierarchical cross-entropy loss and a soft label technique to train these hierarchy-aware models. 
On top of this, \cite{karthiknoERM} develops a post hoc approach using Conditional Risk Minimization (CRM) to calibrate likelihoods during the test phase. Besides, borrowing the idea from neural collapse \cite{papyan2020neuralcollapse}, \cite{liang2023inducing} utilizes an Equiangular Tight Frame to achieve feature alignment across all classes.

Despite significant progress, a critical challenge remains: the current FGVC paradigm struggles with the issue of \textbf{Granularity competition} in feature learning due to the semantic gap between multi-granularity labels (as shown in Fig.\ref{fig:acc_main}). 
Specifically, coarse labels are generally much easier to distinguish than finer ones. 
As a result, if we pay equal attention to all levels, the learning process of the underlying encoder is likely to be dominated by the coarse branches, leading to insufficient learning of detailed information at finer granularities. 
To address this issue, \cite{chang2021yourflamingo} disentangles coarse-level features from fine-grained ones via level-specific classification heads, and also uses the finer-grained features to enhance the performance of coarser-grained predictions. 
\cite{garg2022learningERM} proposes a soft balance strategy to align the learning process of multi-granularity classifications. However, these methods primarily emphasize the negative impacts of coarse-grained learning on finer details, while overlooking the crucial fact that \textbf{coarse-grained information can also enhance fine-grained learning}. 
Consequently, the issue of granularity competition remains \textbf{inadequately considered}.

In this paper, we propose a generic framework called \textit{Bidirectional Logits Tree} (BiLT), designed to fully utilize hierarchical label information. 
Specifically, BiLT builds hierarchy-aware models in a sequential manner rather than a parallel one, where the inputs to the coarse classifier depend on the outputs of the preceding finer-grained model. 
This paradigm improves the priority of learning finer features and also facilitates semantic information flow among multi-granularity labels during training. 
Meanwhile, classification errors at coarser levels could also serve as an auxiliary supervision signal for upstream finer models' updates, reconciling the granularity competition issue end-to-end. 
Last but not least, we recognize that accurately identifying subtle semantic differences between sub-classes within the same granularity is also crucial for achieving promising performance. 
To this end, an Adaptive Intra-Granularity Difference Learning (AIGDL) approach is developed to serve our purpose better.

In summary, the contributions of this paper are as follows:
\begin{itemize}
    \item We propose a novel paradigm called BiLT to construct hierarchy-aware predictors for fine-grained classifications, which can effectively alleviate the granularity competition issue by facilitating the flow of hierarchical semantic information across all granularities. 
    \item An Adaptive Intra-Granularity Difference Learning (AIGDL) method is further developed to serve our strategy better. 
    It empowers BiLT to learn the fine-grained semantic differences of classes within the same granularity, boosting the final performance sharply.
    \item Empirical studies over three widely used benchmark datasets consistently demonstrate the effectiveness of our proposed approach. 
\end{itemize}


\section{Related Work}

\subsubsection{Hierarchical Architecture Methods}
Hierarchical architecture methods aim to design a hierarchy-aware model architecture that leverages the hierarchical relationship. 
\cite{silla2011survey} first introduced hierarchy-aware classifiers that incorporated a well-defined hierarchy, empirically outperforming flat classifiers across various application scenarios. 
Building on this idea, \cite{wu2016learning, bertinetto2020making} proposed connecting classifiers at all levels to a shared feature vector, framing the overall optimization as a multi-task learning problem.
However, \cite{chang2021yourflamingo} later highlighted a potential issue with this method that sharing the same feature vector across multiple granularities can cause granularity competition problems, where coarse-level predictions impede fine-grained feature learning, whereas fine-grained feature learning can facilitate coarse-grained classification.
To avoid this issue, \cite{liang2023inducing} from the perspective of neural collapse, aligned features with class distances by fixing classifier weights to a precomputed Equiangular Tight Frame based on the distance matrix.
However, this method is limited by the requirement that the feature dimension must equal the number of classes, making it difficult to apply to large datasets and certain downstream tasks.
More recently, \cite{jain2024test} proposed a multi-granularity ensemble method, training multiple neural networks at different levels separately, and adjusting fine-grained outputs based on coarse-grained outputs during testing. 
This approach improved Top-1 Accuracy and reduced Mistake Severity in fine-grained classification.
However, this approach requires training multiple networks separately, which is computationally expensive and difficult to scale to datasets with many levels.

\subsubsection{Hierarchical Loss/Cost Methods}
Hierarchical loss methods aim to design loss functions that can effectively leverage the hierarchical relationships between labels.
\cite{bertinetto2020making} proposed a hierarchical cross-entropy (HXE) loss, a conditional probabilistic loss that conditions class probabilities on those of their ancestors.
This approach jointly optimizes the HXE loss across all granularities, framing it as a multi-task learning problem.
In contrast, \cite{karthiknoERM} introduced a cost-sensitive method based on conditional risk minimization, calibrating outputs according to the class-relationship matrix during the test phase to minimize the risk.
As noted earlier, using hard labels for coarse-grained classification can impede fine-grained feature learning.
To address this, \cite{garg2022learningERM} proposed summing the soft labels of subclasses and aligning them with coarse-grained hard labels using Jensen-Shannon divergence, while also aligning subclass features with those of their superclass using a geometric consistency loss.

\subsubsection{Label Embedding Methods}
This approach focuses on uncovering relationships in a unified label semantic space and therefore describes differences between labels by the distance in this space.
For instance, \cite{frome2013devise, xian2016latent} started early attempts to explore a generic algorithm for learning a joint embedding space for images and labels simultaneously. 
Inspired by the strengths of hyperbolic space in modeling hierarchical structures, \cite{liu2020hyperbolic} proposed learning a hyperbolic label space for fine-grained classifications, leading to promising performance. 
In addition, to learn a favorable space, \cite{bertinetto2020making} first initialized relationships between labels based on their lowest common ancestor (LCA) height, and proposed a soft label method for precise fine-tuning. 
Taking a step further, \cite{zhang2021delving} developed an online label smoothing strategy to adaptively learn the label embedding during training, while \cite{collins2022eliciting} proposed a crowdsourced soft label method for individual images.
Additionally, label smoothing plays a significant role in various tasks, including but not limited to knowledge distillation\cite{yuan2020revisiting, park2023acls, han2024aucseg}, image generation\cite{zhang2023diffsmooth}, weakly supervised learning\cite{gong2024does, wei2022smooth}, and trustworthy machine learning\cite{qin2021improving}. 
Overall, label smoothing is a widely used technique to enhance model generalizability and performance\cite{muller2019does}.
Nevertheless, these methods often fail to utilize prior knowledge of inter-label distances. 
They heavily rely on predefined class relationships, or even solely depend on heuristic strategies for label embedding adjustment, which results in suboptimal performance and slow convergence.


\begin{figure*}[!t]
    \centering
    \includegraphics[width=1.95\columnwidth]{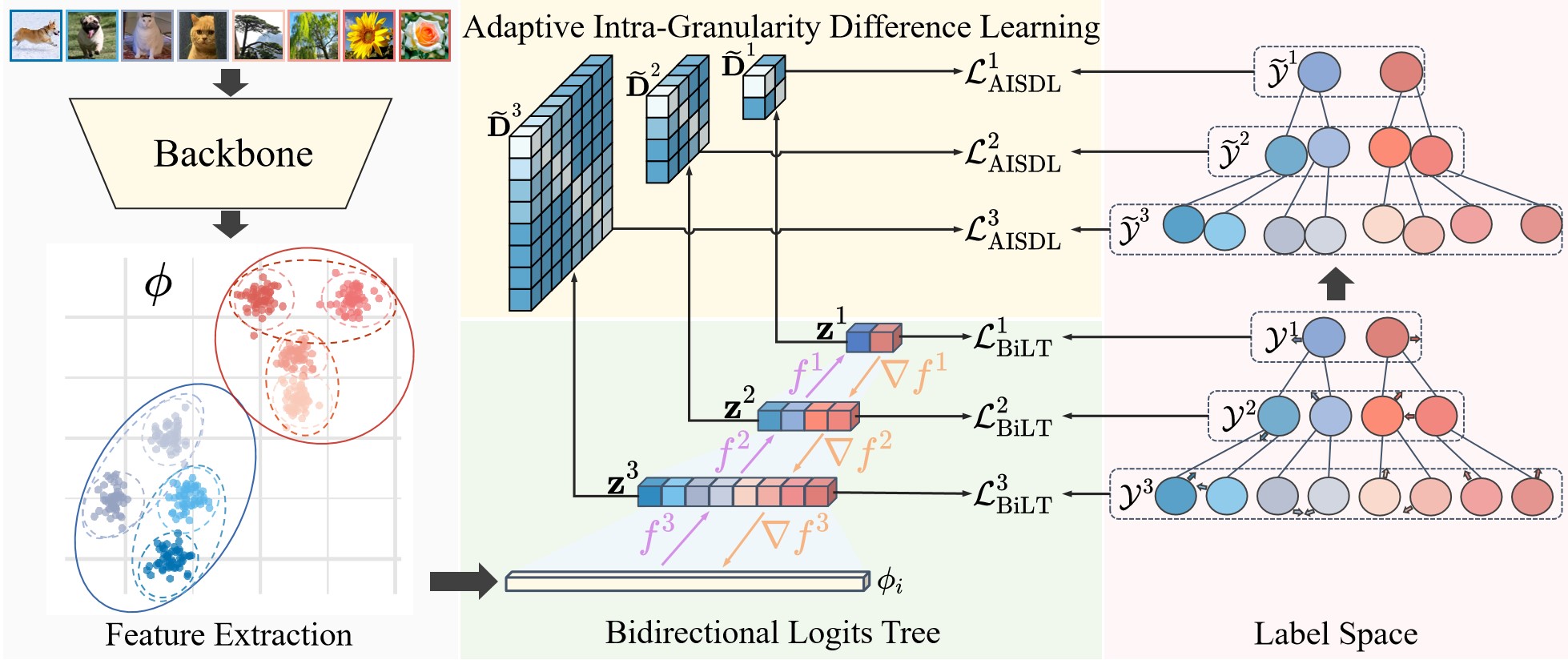}
        \caption{The overall framework of our method. In the \textbf{forward} phase of the Bidirectional Logits Tree (BiLT), coarse-grained logits are derived from fine-grained counterparts, while the gradients from coarse-grained classifiers influence fine-grained classifiers and feature learning in the \textbf{backward} phase. 
    Simultaneously, Adaptive Intra-Granularity Difference Learning (AIGDL) adjusts the output of BiLT and supervision by learning differences between categories within the same granularity.
    } 
    \label{fig:Pipline}
\end{figure*}

\section{Problem Definition}
Let $\mathcal{X}$ and $\mathcal{Y}$ be the input space and label space, respectively. 
Under the context of fine-grained visual classifications (FGVC) \cite{liu2024democratizing, wang2024transhp, du2021progressive}, the label space $\mathcal{Y}$ could be formulated as a hierarchy label tree with $H+1$ levels, where each node corresponds to a specific class, and each edge contains high-level semantic relationships between classes. 
Typically, for $h \in \{0, 1, 2, \ldots, H\}$, the root node ($h=0$) represents the super-class of all classes, and the label becomes finer and finer as the level $h$ increases in the tree. 
Subsequently, let $C^h$ be the number of classes at the $h$-th level, the label space could be expressed as $\mathcal{Y} = \mathop{\cup}\limits_{h=0}^{H} \mathcal{Y}_h$, where $\mathcal{Y}_h = \{0, 1, \ldots, C^h-1\}$ and we have $C^0 = 1 < C^1 \leq C^2 \leq \cdots \leq C^H$. 
Finally, suppose that there are $N$ training samples, denoted as $\mathcal{D} = \{(\mathbf{x}_i, \{y^h_i\}_{h=1}^{H} )\}_{i=1}^{N}$, where $\mathbf{x}_i \in \mathcal{X}$ is an input image and $y^h_i \in \mathcal{Y}_h$ corresponds to its ground-truth label at the $h$-th level. 
The primary concern of this paper is to train a well-performed classifier that can accurately predict the fine-grained class for each image while ensuring that the predictions are close to the ground truth when it makes mistake.

\begin{figure}[!t]
    \centering
    \subfloat[Sharing Same Features]{
        \includegraphics[width=0.46\linewidth]{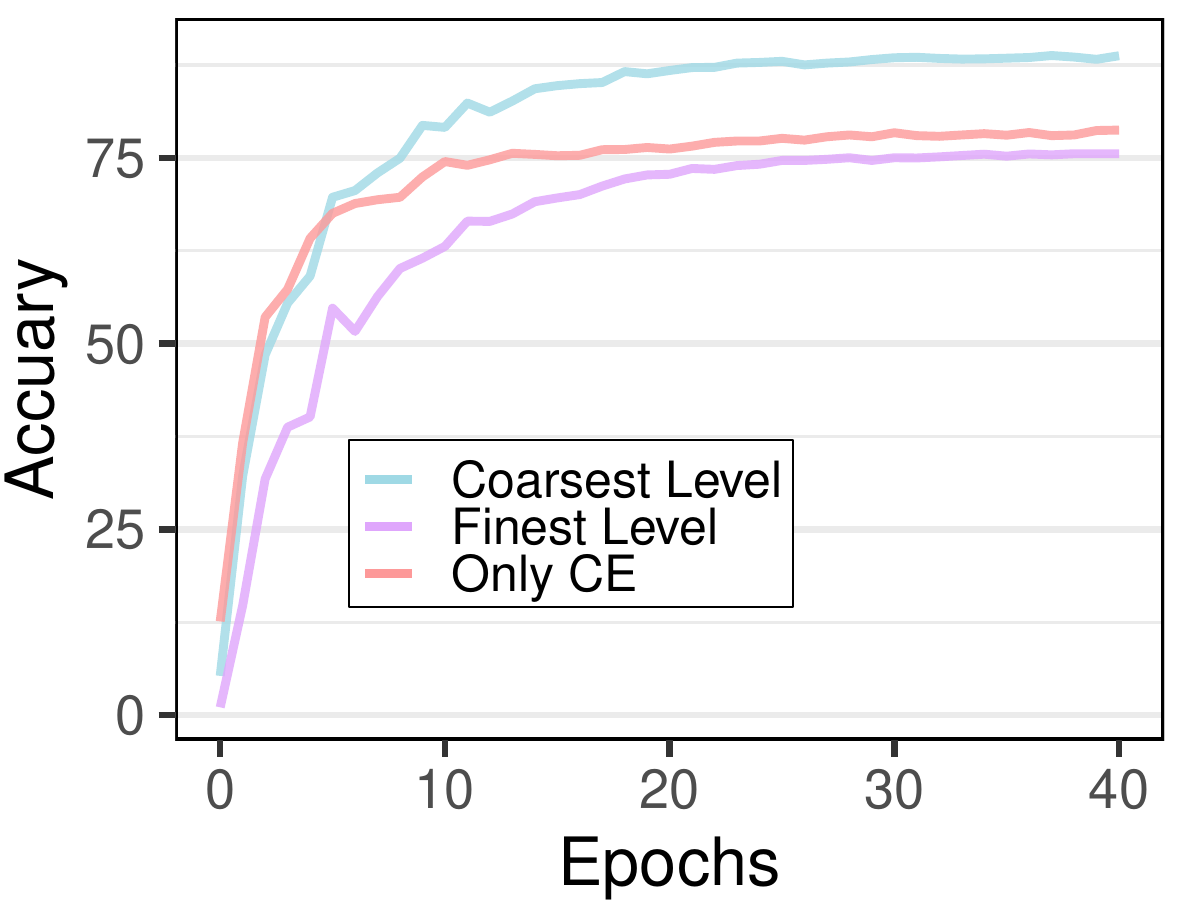}
        \label{fig:share_same_feature_acc}
    }
    \subfloat[BiLT]{
        \includegraphics[width=0.46\linewidth]{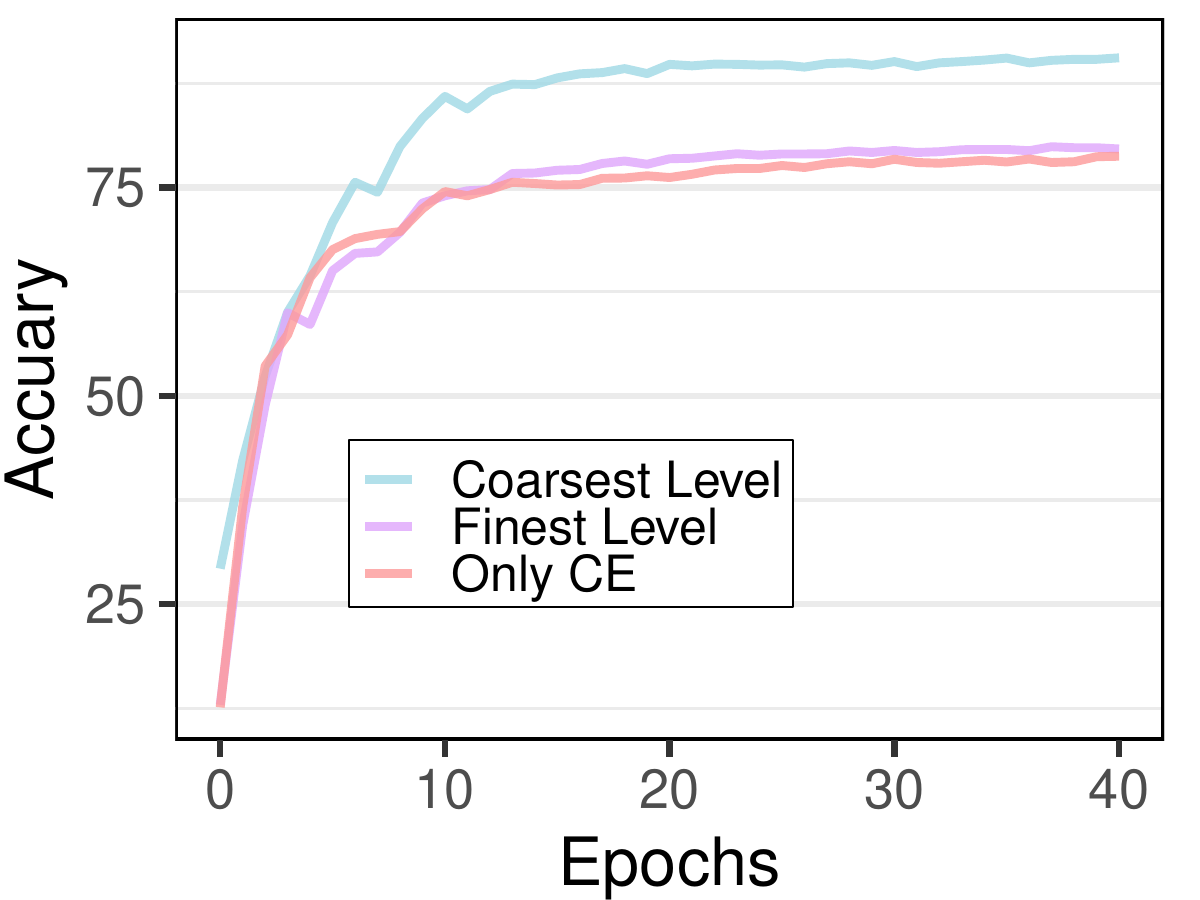}
        \label{fig:BiLT_acc}
    }
    \caption{Comparison of convergence speed of Sharing Same Features and BiLT}
    \label{fig:acc_main}
\end{figure}

\section{Methodology}
In this section, we will first reveal the fundamental limitations of current fine-grained classification methods, i.e., \textit{the granularity competition issue}. 
To address this, we propose a generic framework called Bidirectional Logits Tree (BiLT). 
Fig. \ref{fig:Pipline} presents the overall pipeline of our proposed BiLT. 
On top of this, a novel \textit{Adaptive Intra-Granularity Difference Learning} is developed to exploit the fine-grained semantics among all classes sufficiently. 
Further details on BiLT will be discussed in the following.

\subsection{Granularity Competition Impedes Fine-grained Learning}
\begin{figure}[!t]
    \centering
    \includegraphics[width=0.9\columnwidth]{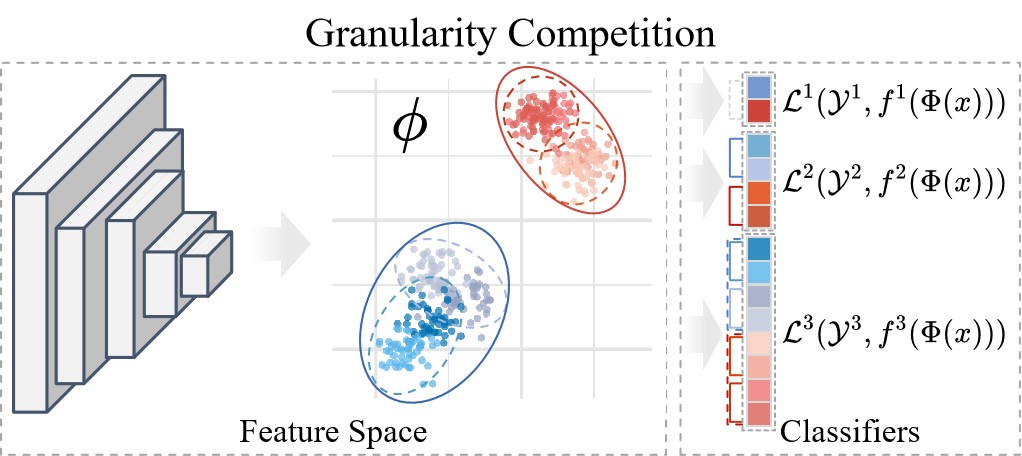}
    \caption{An illustrative example demonstrates the granularity competition problem, where the model prioritizes coarse-grained learning at Level 1 and Level 2, thereby rendering the fine-grained features at Level 3 difficult to distinguish.} 
    \label{fig:Confrontation}
\end{figure}

We start with our discussions by reviewing the traditional FGVC paradigms \cite{wang2015multiple} using the hierarchy label tree. 
Briefly speaking, given an image $x$ with its label sets $\{y^h\}_{h=1}^{H}$, conventional FGVC methods using the hierarchy label tree first adopt a base encoder $\Phi$ to extract the semantic features of each sample, i.e., $\Phi(x)$. 
On this basis, an independent classifier (denoted as $f^h)$) will be constructed for each label level $h$ whose goal is to make an accurate prediction $\hat{y}^h$ as much as possible at its own level. 
In order to learn these classifiers effectively, current studies \cite{wang2024transhp, garg2022learningERM, chang2021yourflamingo}
usually consider the following multi-task optimization problem:
\begin{equation}\label{eq1}
    \min\limits_{\Phi, f^1, \dotsm f^H} \mathop{\mathbb{E}}\limits_{\mathcal{D}}\left[
    \sum_{h=1}^H \lambda_h \cdot \mathcal{L}^h(\mathcal{Y}^h, f^h(\Phi(x)))\right],
\end{equation}
where $\lambda_h$ is a tunable parameter, and $\mathcal{L}^h$ represents the classification error (such as cross-entropy loss) at $h$-th level in the hierarchy label tree. 
Fig. \ref{fig:Confrontation} presents a toy example ($H=3$) to instantiate the mainstream learning paradigm of FGVC.

Despite great success, in this paper, we argue that current FGVC paradigms using the hierarchy label tree, suffer from \textbf{\textit{Granularity competition}} issues, leading to limited performance. 
According to Eq.\ref{eq1}, it is apparent that coarse-grained levels are inherently simpler to learn than fine-grains, so the learning process of the base encoder $\Phi$  will be almost dominated by the shallow level (i.e., coarse labels). 
Fig. \ref{fig:acc_main} examines this phenomenon on the real-world benchmark dataset FGVC-Aircraft, showing that \textbf{methods sharing the same features at fine-grained levels converge more slowly and yield lower accuracy} compared to only fine-grained level trained exclusively with standard cross-entropy loss (Only CE). 
As a result, the unique features extracted by $\Phi$ inevitably collapse, where the detailed information for fine-grained feature learns insufficiently \cite{chang2021yourflamingo}. 
In addition, current paradigms have not fully explored the semantic relationships between classes at different granularities, which further impedes learning fine-grained features for accurate predictions.

\subsection{Bidirectional Logits Tree for Granularity Reconcilement}
According to the above discussions, the fundamental limitations of previous FGVC literature \cite{bertinetto2020making, chang2021yourflamingo, garg2022learningERM} come from the fact that all classifiers at different granularities built on unique base encoders. 
We propose a novel generic framework called Bidirectional Logits Tree (BiLT) to address this. 
The principle of BiLT is to develop coarse-level classifiers on top of its previous finer classifier outputs instead of constructing a set of classifiers separately, as shown in Fig. \ref{fig:Pipline}. 

Specifically, let $\mathbf{\phi}_i:=\Phi(x_i)$ be the feature vector for simplicity. BiLT first constructs the finest-level classifier $f^H$, which directly takes $\mathbf{\phi}_i$ as the input, i.e., 
$$\mathbf{z}^H_i := f^H(\mathbf{\phi}_i),$$
where $\mathbf{z}^H_i \in \mathbb{R}^{C_H}$ is the output logits at level $H$, and $z^H_{ij}, j \in [C_H]$ indicates the probability of image $x_i$ belonging to the class $j$. Subsequently, for the class-level $h \in \{1,\dots, H-1\}$, its corresponding classifier $f^{h}$ is developed following the previous $h-1$ outputs:
\begin{equation*}
    \mathbf{z}^{h}_i := f^{h}(g^{h}(\mathbf{z}^{h+1}_i)),
\end{equation*}
where $\mathbf{z}^{h}_i \in \mathbb{R}^{C^h}$ and $g^{h}$ is a trainable transformation layer to explore nonlinear relationships between different levels. 
In this paper, $g^{h}$ is implemented by a sequential module, including batch normalization, linear transformation, batch normalization, and ELU activation \cite{liang2023inducing}. 

We still adopt a similar manner as Eq.1 to learn BiLT. Here the loss for each sample $x_i$ at $h$-th level is defined as:
\begin{equation*}
    \mathcal{L}^h_{\mathrm{BiLT}}(\mathcal{Y}^h, f^h) = \mathbf{y}^h_i \log{(\mathrm{softmax}(\mathbf{z}^h_i))},
\end{equation*}
where $\mathbf{y}^h_i$ is a one-hot encoding with the $y_i^h$-th indices as $1$. 

Intuitively, the learning process of our proposed BiLT \textit{boosts a bidirectional semantic information flow across all granularities} properly, as shown in Fig. \ref{fig:Pipline}.
First of all, during the \textbf{forward propagation}, the finer outputs serve as the input features for the coarse-level model recursively, enabling cross-level information explorations. 
This to some extent \textbf{promotes the learning priority of finer features} in the base encoder $\Phi$ such that more discriminative features will be abstracted for finer classifications. 
Meanwhile, during the \textbf{backward propagation}, each classifier is not only optimized by the classification risk at the current level but also supervised by those coarser levels. 
In this way, classifiers at different granularities will interact with each other, leverage hierarchical information effectively, and thus reconcile the granularity competition problem.


\begin{figure}[!t]
    \centering
    \includegraphics[width=0.92\columnwidth]{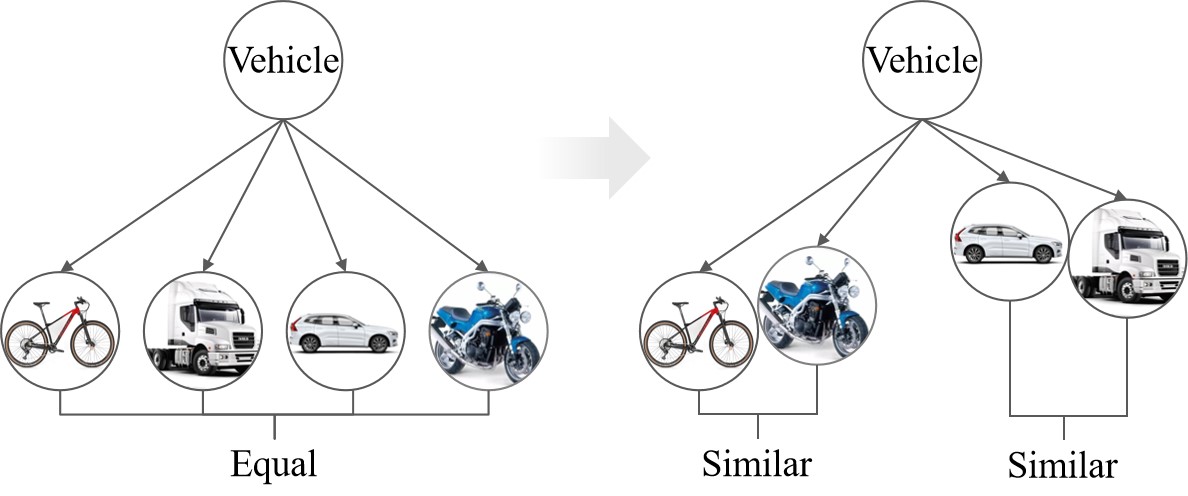}
    \caption{Predefined label trees struggle to articulate differences amongst classes at the same hierarchical level, and our method aims to learn disparities among classes and apply relevant corrections accordingly.} 
    \label{fig:label_tree}
\end{figure}

\subsection{Adaptive Intra-Granularity Difference Learning}
So far, we have explored how to excavate fine-grained class relationships effectively at different levels. 
However, mining the semantic information between labels involved \textbf{at the same granularity} is also essential for promising performance. 
Current studies \cite{garg2022learningERM,liang2023inducing} generally introduce a cost matrix defined on LCA (Least Common Ancestor) to calibrate the final predictions. 
Based on the fact that there are often obscure semantic differences between sub-classes at the same granularity, \textbf{such a strategy} considering intra-granularity classes equally might \textbf{struggle to} articulate them. Fig. \ref{fig:label_tree} provides an example to illustrate it. 
We know that both ``bicycles", ``motorcycles", ``trucks" and ``car" belong to ``Vehicle". 
Yet, in the feature space, it is obvious that the distance between ``bicycles" and ``motorcycles" should be closer than ``bicycles" and ``trucks" since their appearance is highly similar. 
Apparently, this cannot be achieved through simple LCA-based methods.

We propose a novel Adaptive Intra-Granularity Difference Learning (AIGDL) method to remedy this. 
Concretely, let $\mathbf{{p}}^h_i=\mathrm{softmax}(\mathbf{z}^h_i)$ be the prediction probability and $\mathbf{D}^h \in \mathbb{R}^{C^h \times C^h}$ be the intra-granularity class distance matrix at level $h$, where $D^h_{ij}$ represents the $\textnormal{LCA}$ distance between class $y^h_i$ and class $y^h_j$. 
We first follow the widely adopted Condition Risk Minimization (CRM) paradigm \cite{garg2022learningERM} to calibrate:
\begin{align*}
    \mathop{\mathrm{argmin}}\limits_k & R(\mathbf{y}^h_i = k|\mathbf{x}_i) = 
    \mathop{\mathrm{argmin}}\limits_k \sum_{j=1}^{C^h} \mathbf{D}^h_{k,j} \cdot {p}^h(\mathbf{y}^h_i = k|\mathbf{x}_i).
\end{align*}

Compared with the typical likelihood maximization, this CRM adjustment guarantees us to select the Bayes optimal prediction $R(\mathbf{y}^h_i=k|\mathbf{x}_i)$ resulting in the lowest possible overall cost \cite{Duda1974PatternCA}.

On top of $\mathbf{D}^h$, a learnable intra-granularity distance matrix $\mathbf{\Delta}^h \in \mathbb{R}^{C^h \times C^h}$ is further introduced to capture the nuanced semantic relationships between classes. Thereafter, the decision rule is augmented as follows:
\begin{align*}
    &\mathop{\mathrm{argmin}}\limits_k  R(\mathbf{y}^h_i|\mathbf{x}_i) \\
    = &\mathop{\mathrm{argmin}}\limits_k  \sum_{j=1}^{C^h} (D^h_{k,j} - \beta \cdot \Delta^h_{k,j}) \cdot {p}^h(\mathbf{y}^h_i = k|\mathbf{x}_i) \\ 
    = &\mathop{\mathrm{argmax}}\limits_k  \sum_{j=1}^{C^h} (\beta \cdot \Delta^h_{k,j} - D^h_{k,j}) \cdot {p}^h(\mathbf{y}^h_i = k|\mathbf{x}_i) \\
    = &\mathop{\mathrm{argmax}}\limits_k  \sum_{j=1}^{C^h} \widetilde{D}^h_{k,j} \cdot {p}^h(\mathbf{y}^h_i = k|\mathbf{x}_i),
\end{align*}
where $\widetilde{\mathbf{D}}^h=\beta \cdot \mathbf{\Delta}^h - \mathbf{D}^h$ and $\beta$ signifies the weight of the learnable matrix $\mathbf{\Delta}^h$. 
Meanwhile, to avoid self-cost, the diagonal entries of $\mathbf{\Delta}^h$ (i.e., $\mathbf{\Delta}^h_{ii}$) are masked as zeros and then each row of $\mathbf{\Delta}^h$ undergoes an L2 normalization.

To effectively learn intra-granularity differences, we thus introduce a label smoothing strategy \cite{muller2019does} for training. Different from previous studies \cite{bertinetto2020making}, as the goal is to pursue a model with better mistake, we propose to employ the above class-wise matrix to label smoothing explicitly. 
Specifically, the label smoothing technique for AIGDL could be expressed as:
\begin{equation*}
    {\widetilde{\mathbf{y}}^h_i}=( 1 - \epsilon ){\mathbf{y}^h_i} + \epsilon \cdot \frac{\exp{(\gamma(\beta \cdot \mathbf{\Delta}_{y^h_i}^h - \mathbf{D}_{y^h_i}^h))}}{\sum_{j=1}^{C^h} \exp{(\gamma(\beta \cdot \mathbf{\Delta}_{y^h_i, j}^h - \mathbf{D}_{y^h_i, j}^h))}},
\end{equation*}
where $\epsilon$ is the smoothing factor, $\gamma$ is the temperature parameter, and $\mathbf{D}_{k}^h$ and $\mathbf{\Delta}_{k}^h$ are the $k$-th row of $\mathbf{D}^h$ and $\mathbf{\Delta}^h$ respectively. In this way, the semantic similarities among these subclasses can be implicitly reflected at the label level, ensuring consistency in predictions. Moreover, it is to be noted that the above smoothing method is a general version of the Soft-Label method \cite{bertinetto2020making} by setting $\epsilon=1$ and $\beta=0$. 

Similarly, the loss function of AIGDL for each level is defined as:
\begin{equation*}
    \mathcal{L}^h_{\textnormal{AIGDL}} = \frac{1}{N} \sum_{i=1}^{N} {\widetilde{\mathbf{y}}^h_i} \log{(\mathrm{softmax}(\widetilde{\mathbf{D}}^h\mathbf{p}^h_i))}.
\end{equation*}

\subsection{Overall Optimization Objective}
Combining all the above components, the overall optimization objective is defined as:
\begin{align*}
    &\mathcal{L} = \sum_h \lambda_h \cdot (\mathcal{L}^h_{\textnormal{BiLT}} + \mathcal{L}^h_{\textnormal{AIGDL}}) \\ 
    &= \sum_h \frac{\lambda_h}{N} \sum_{i=1}^{N} (\mathbf{y}^h_i \log{(\mathbf{{p}}^h_i)} + {\widetilde{\mathbf{y}}^h_i} \log{(\mathrm{softmax}(\widetilde{\mathbf{D}}^h\mathbf{p}^h_i))}),
\end{align*}
where $\lambda_h = \exp(\alpha \cdot (h-H))$ is the weight for each level. 
As the hierarchical level becomes coarser, its contribution to fine-grained feature learning diminishes. 
So the corresponding loss weight $\lambda_h$ decreases as the hierarchical level becomes coarser.


\begin{table*}[!ht]
    \centering
    \resizebox{0.95\textwidth}{!}{
    \begin{tabular}{c|c||ccc||c}
    \toprule
    Method & Mistake Severity($\downarrow$) & Hier Dist@1($\downarrow$) & Hier Dist@5($\downarrow$) & Hier Dist@20($\downarrow$) & Top-1 Accuracy($\uparrow$) \\
    \midrule
    Cross-Entropy & \cellcolor[rgb]{ .965,  .980,  .992}{2.12 +/- 0.0288}           & \cellcolor[rgb]{ .965,  .980,  .988}{0.44 +/- 0.0188}           & \cellcolor[rgb]{ .999,  .999,  .999}{2.10 +/- 0.0053}           & \cellcolor[rgb]{ .999,  .999,  .999}{2.67 +/- 0.0022}           & \cellcolor[rgb]{ .961,  .980,  .992}{79.35 +/- 0.7021} \\
    HXE           & \cellcolor[rgb]{ .949,  .973,  .988}{2.04 +/- 0.0074}           & \cellcolor[rgb]{ .961,  .976,  .988}{0.43 +/- 0.0283}           & \cellcolor[rgb]{ .999,  .999,  .999}{1.96 +/- 0.0119}           & \cellcolor[rgb]{ .999,  .999,  .999}{2.60 +/- 0.0085}           & \cellcolor[rgb]{ .965,  .980,  .992}{78.75 +/- 0.9481} \\
    Soft-Labels   & \cellcolor[rgb]{ .961,  .976,  .988}{2.10 +/- 0.0124}           & \cellcolor[rgb]{ .984,  .992,  .996}{0.49 +/- 0.0223}           & \cellcolor[rgb]{ .999,  .999,  .999}{2.07 +/- 0.0149}           & \cellcolor[rgb]{ .999,  .999,  .999}{2.65 +/- 0.0058}           & \cellcolor[rgb]{ .973,  .984,  .996}{77.59 +/- 0.9698} \\
    Flamingo      & \cellcolor[rgb]{ .961,  .976,  .988}{2.10 +/- 0.0352}           & \cellcolor[rgb]{ .949,  .973,  .988}{0.40 +/- 0.0116}           & \cellcolor[rgb]{ .999,  .999,  .999}{2.06 +/- 0.0099}           & \cellcolor[rgb]{ .999,  .999,  .999}{2.65 +/- 0.0043}           & \cellcolor[rgb]{ .894,  .937,  .976}\underline{80.72 +/- 0.5849} \\
    CRM           & \cellcolor[rgb]{ .957,  .976,  .988}{2.08 +/- 0.0366}           & \cellcolor[rgb]{ .957,  .976,  .988}{0.42 +/- 0.0163}           & \cellcolor[rgb]{ .878,  .929,  .969}\underline{1.74 +/- 0.0053} & \cellcolor[rgb]{ .741,  .843,  .933}\underline{2.44 +/- 0.0018} & \cellcolor[rgb]{ .957,  .976,  .992}{79.57 +/- 0.5880} \\
    HAF           & \cellcolor[rgb]{ .999,  .999,  .999}{2.53 +/- 0.0610}           & \cellcolor[rgb]{ .999,  .999,  .999}{0.67 +/- 0.0295}           & \cellcolor[rgb]{ .999,  .999,  .999}{2.10 +/- 0.0063}           & \cellcolor[rgb]{ .999,  .999,  .999}{2.61 +/- 0.0028}           & \cellcolor[rgb]{ .999,  .999,  .999}{73.68 +/- 1.2166} \\
    HAFrame       & \cellcolor[rgb]{ .792,  .875,  .945}\underline{2.01 +/- 0.0103} & \cellcolor[rgb]{ .843,  .906,  .961}\underline{0.39 +/- 0.0162} & \cellcolor[rgb]{ .949,  .973,  .988}{1.75 +/- 0.0060}           & \cellcolor[rgb]{ .949,  .973,  .988}{2.45 +/- 0.0015}           & \cellcolor[rgb]{ .949,  .973,  .988}{80.52 +/- 0.8375} \\
    HiE           & \cellcolor[rgb]{ .953,  .973,  .988}{2.06 +/- 0.0279}           & \cellcolor[rgb]{ .965,  .980,  .988}{0.44 +/- 0.0146}           & \cellcolor[rgb]{ .969,  .984,  .992}{1.82 +/- 0.0028}           & \cellcolor[rgb]{ .953,  .973,  .988}{2.46 +/- 0.0013}           & \cellcolor[rgb]{ .965,  .980,  .992}{78.66 +/- 0.9660} \\
    Ours          & \cellcolor[rgb]{ .741,  .843,  .933}\textbf{2.00 +/- 0.0220}    & \cellcolor[rgb]{ .741,  .843,  .933}\textbf{0.38 +/- 0.0097}    & \cellcolor[rgb]{ .741,  .843,  .933}\textbf{1.72 +/- 0.0032}    & \cellcolor[rgb]{ .741,  .843,  .933}\textbf{2.44 +/- 0.0016}    & \cellcolor[rgb]{ .741,  .843,  .933}\textbf{81.23 +/- 0.5820} \\
    \bottomrule
    \end{tabular}
    }
    \caption{Performance comparisons on the FGVC-Aircraft dataset with different metrics. The first and second best results are highlighted with \textbf{bold} text and \underline{underline}, respectively.}
    \label{tab:fgvc-air}
\end{table*}

\begin{table*}[t]
    \centering
    \resizebox{0.95\textwidth}{!}{
    \begin{tabular}{c|c||ccc||c}
        \toprule
        Method & Mistake Severity($\downarrow$) & Hier Dist@1($\downarrow$) & Hier Dist@5($\downarrow$) & Hier Dist@20($\downarrow$) & Top-1 Accuracy($\uparrow$)\\
        \midrule
        Cross-Entropy & \cellcolor[rgb]{ .992,  .976,  .965}{2.35 +/- 0.0225}           & \cellcolor[rgb]{ .984,  .937,  .894}{0.52 +/- 0.0076}           & \cellcolor[rgb]{ .999,  .999,  .999}{2.25 +/- 0.0146}           & \cellcolor[rgb]{ .999,  .999,  .999}{3.18 +/- 0.0079}        & \cellcolor[rgb]{ .984,  .922,  .863} \underline{77.85 +/- 0.2699} \\
        HXE           & \cellcolor[rgb]{ .999,  .999,  .999}{2.40 +/- 0.0137}           & \cellcolor[rgb]{ .999,  .999,  .999}{0.62 +/- 0.0205}           & \cellcolor[rgb]{ .996,  .984,  .973}{2.05 +/- 0.0082}           & \cellcolor[rgb]{ .996,  .984,  .976}{3.02 +/- 0.0171}        & \cellcolor[rgb]{ .999,  .999,  .999} 72.76 +/- 0.6816 \\
        Soft-Labels   & \cellcolor[rgb]{ .992,  .969,  .949}{2.33 +/- 0.0270}           & \cellcolor[rgb]{ .999,  .999,  .999}{0.62 +/- 0.0144}           & \cellcolor[rgb]{ .984,  .929,  .882}{1.29 +/- 0.0056}           & \cellcolor[rgb]{ .992,  .961,  .937}{2.73 +/- 0.0122}        & \cellcolor[rgb]{ .996,  .976,  .961} 74.34 +/- 0.4588 \\
        Flamingo      & \cellcolor[rgb]{ .992,  .965,  .941}{2.32 +/- 0.0186}           & \cellcolor[rgb]{ .984,  .933,  .886}{0.51 +/- 0.0094}           & \cellcolor[rgb]{ .996,  .984,  .976}{2.07 +/- 0.0214}           & \cellcolor[rgb]{ .996,  .988,  .984}{3.08 +/- 0.0077}        & \cellcolor[rgb]{ .984,  .922,  .867} 77.83 +/- 0.2942 \\
        CRM           & \cellcolor[rgb]{ .992,  .961,  .933}{2.31 +/- 0.0229}           & \cellcolor[rgb]{ .984,  .933,  .886}{0.51 +/- 0.0019}           & \cellcolor[rgb]{ .984,  .922,  .867}{1.15 +/- 0.0037}           & \cellcolor[rgb]{ .984,  .922,  .867}{2.20 +/- 0.0034}        & \cellcolor[rgb]{ .988,  .925,  .871} 77.80 +/- 0.2462 \\
        HAF           & \cellcolor[rgb]{ .984,  .933,  .886}{2.24 +/- 0.0177}           & \cellcolor[rgb]{ .984,  .925,  .875}{0.50 +/- 0.0075}           & \cellcolor[rgb]{ .984,  .937,  .898}{1.42 +/- 0.0057}           & \cellcolor[rgb]{ .988,  .953,  .925}{2.64 +/- 0.0037}        & \cellcolor[rgb]{ .988,  .929,  .878} 77.51 +/- 0.3091 \\
        HAFrame       & \cellcolor[rgb]{ .984,  .922,  .867}{2.21 +/- 0.0108}           & \cellcolor[rgb]{ .984,  .922,  .867}{0.49 +/- 0.0066}           & \cellcolor[rgb]{ .976,  .851,  .757}\underline{1.11 +/- 0.0018} & \cellcolor[rgb]{ .976,  .839,  .741}\underline{2.18 +/- 0.0013}& \cellcolor[rgb]{ .988,  .925,  .871} 77.71 +/- 0.2319 \\
        HiE           & \cellcolor[rgb]{ .980,  .890,  .820}\underline{2.20 +/- 0.0232} & \cellcolor[rgb]{ .976,  .800,  .678}\textbf{0.47 +/- 0.0056}    & \cellcolor[rgb]{ .984,  .922,  .871}{1.19 +/- 0.0031}           & \cellcolor[rgb]{ .992,  .973,  .953}{2.86 +/- 0.0097}        & \cellcolor[rgb]{ .976,  .800,  .678} \textbf{78.63 +/- 0.3101} \\
        Ours          & \cellcolor[rgb]{ .976,  .800,  .678}\textbf{2.17 +/- 0.0197}    & \cellcolor[rgb]{ .980,  .859,  .773}\underline{0.48 +/- 0.0106} & \cellcolor[rgb]{ .976,  .800,  .678}\textbf{1.08 +/- 0.0057}    & \cellcolor[rgb]{ .976,  .800,  .678}\textbf{2.17 +/- 0.0030} & \cellcolor[rgb]{ .988,  .925,  .875} 77.77 +/- 0.4140 \\
        \bottomrule
    \end{tabular}%
    }
    \caption{Performance comparisons on the CIFAR-100 dataset with different metrics. The first and second best results are highlighted with \textbf{bold} text and \underline{underline}, respectively.}
    \label{tab:cifar-100}%
\end{table*}%

\section{Experiments}
\subsection{Experiments Setup}
\subsubsection{Datasets.}
We evaluate methods on four datasets: FGVC-Aircraft \cite{maji2013fine}, CIFAR-100 \cite{krizhevsky2009learning}, iNaturalist2019 \cite{van2018inaturalist} and tieredImageNet-H\cite{ren2018meta}, all of which have been used in previous studies\cite{liang2023inducing, garg2022learningERM, bertinetto2020making}. 
Dataset statistics and split settings are provided in the Appendix.



\subsubsection{Competitors.}
We compare our method against the following competitors: 
\textbf{HXE}\cite{bertinetto2020making},
\textbf{Soft-Labels}\cite{bertinetto2020making}, 
\textbf{Flamingo}\cite{chang2021yourflamingo},
\textbf{CRM}\cite{karthiknoERM}, 
\textbf{HAF}\cite{garg2022learningERM}, 
\textbf{HAFrame}\cite{liang2023inducing}, 
and \textbf{HiE}\cite{jain2024test}. 
We also include the vanilla cross-entropy loss as a baseline, denoted as \textbf{Cross-Entropy}. Note that, all competitors are implemented following the guidelines suggested by the corresponding paper.

\subsection{Performance Comparisons}

\begin{figure*}[!t]
    \centering
    \subfloat[FGVC-Aircraft]{
        \includegraphics[width=0.32\linewidth]{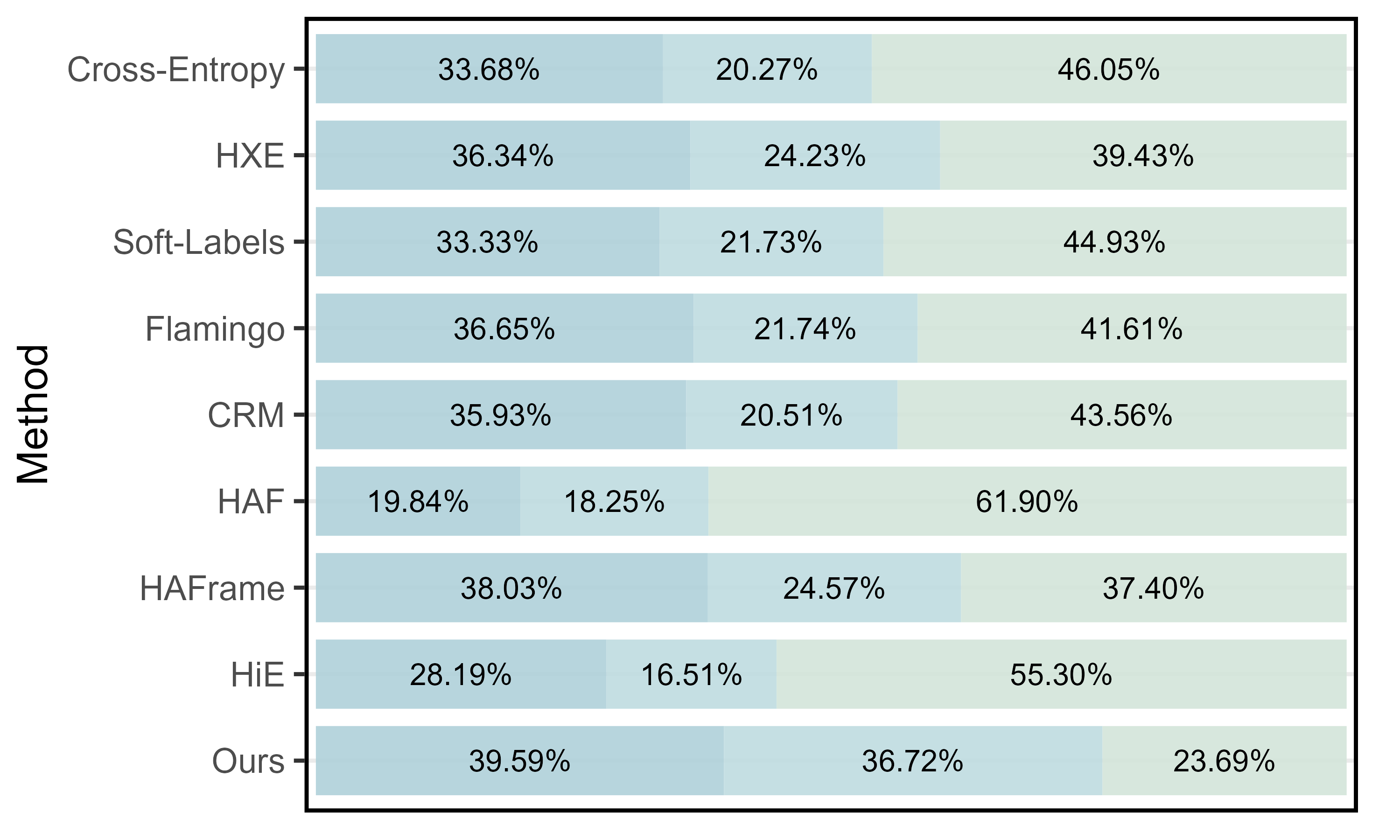}
        \label{fig:fgvc_dist}
    }
    \subfloat[CIFAR-100]{
        \includegraphics[width=0.32\linewidth]{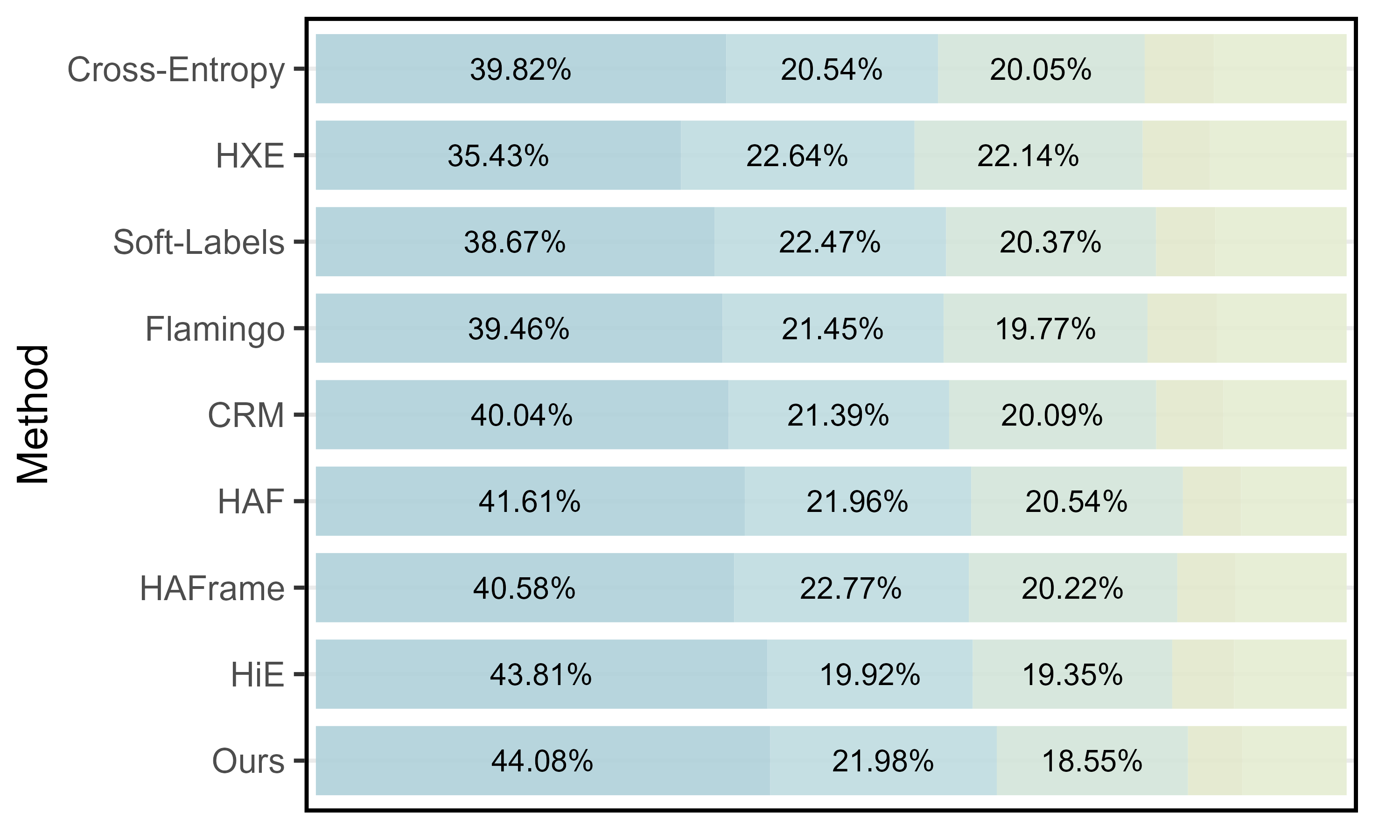}
        \label{fig:cifar100_dist}
    }
    \subfloat[iNaturalist2019]{
        \includegraphics[width=0.32\linewidth]{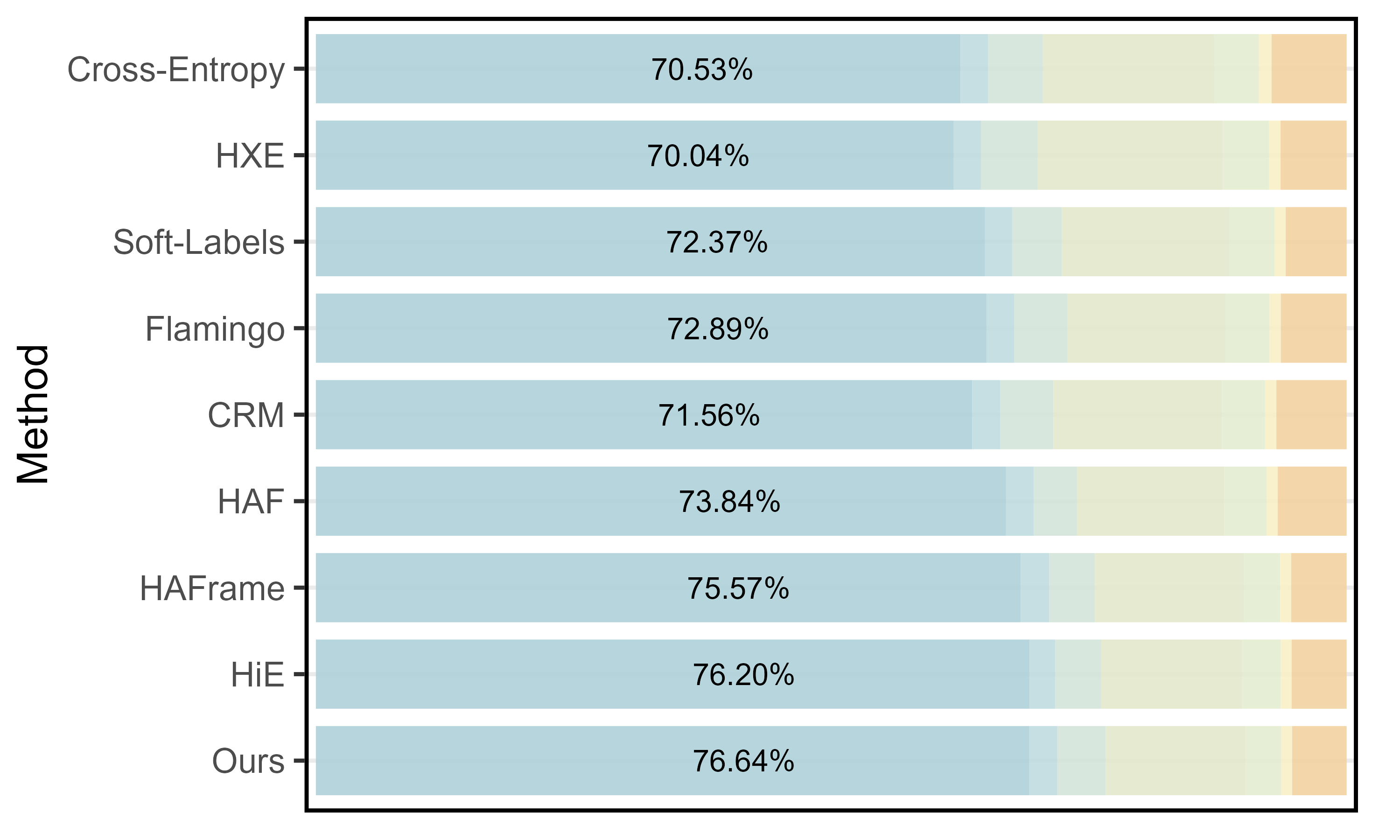}
        \label{fig:inat19_dist}
    }
    \caption{The probability of mistakes made by our method and competitors across the three datasets. 
            In each subfigure, different colors represent varying LCA distances, increasing from left to right. 
            The length of each bar indicates the severity of the mistake, while the values on the bars show the probability of the model making a mistake at a specific LCA distance.}
    \label{fig:hierdist_main_figure}
\end{figure*}

\subsubsection{Overall Performance.}
Tab. \ref{tab:fgvc-air} and Tab. \ref{tab:cifar-100} present the detailed performance of the FGVC-Aircraft and CIFAR-100 datasets, respectively.
Detailed performance results for the iNaturalist2019 dataset are provided in the Appendix.
Notably, all metrics are correlated to log-scaled LCA distance except for Top-1 Accuracy \cite{karthiknoERM}. 
According to the reported results, we can draw the following remarks: 
In most cases, our proposed BiLT could outperform all competitors at all metrics, except for the Top-1 Accuracy and Hier Dist@1 on the CIFAR-100 and iNaturalist2019 datasets. 
Even in failure cases, the performance of BiLT is still comparable. 
This speaks to the efficacy of our approach.

\subsubsection{Mistake Severity Distribution.} 
Fig. \ref{fig:hierdist_main_figure} shows the mistake severity distribution for our method and the competitors. 
Specifically, here we report the performance comparisons at the finest granularity by examining the LCA distances between wrong and ground truth classes. 
It is obvious that, when misclassifying an example, our proposed BiLT could make a more reasonable prediction. 
As shown in Fig. \ref{fig:fgvc_dist}, when the model makes a mistake, our method has a 39.59\% probability that the LCA distance between the prediction and the ground truth is 1. 
Additionally, there is a 76.31\% probability (39.59\% + 36.72\%) that the LCA distance is less than or equal to 2, 
resulting in a significantly lower proportion of severe mistakes compared to those of competitors.


\begin{figure}[!t]
    \centering
    \includegraphics[width=0.72\columnwidth]{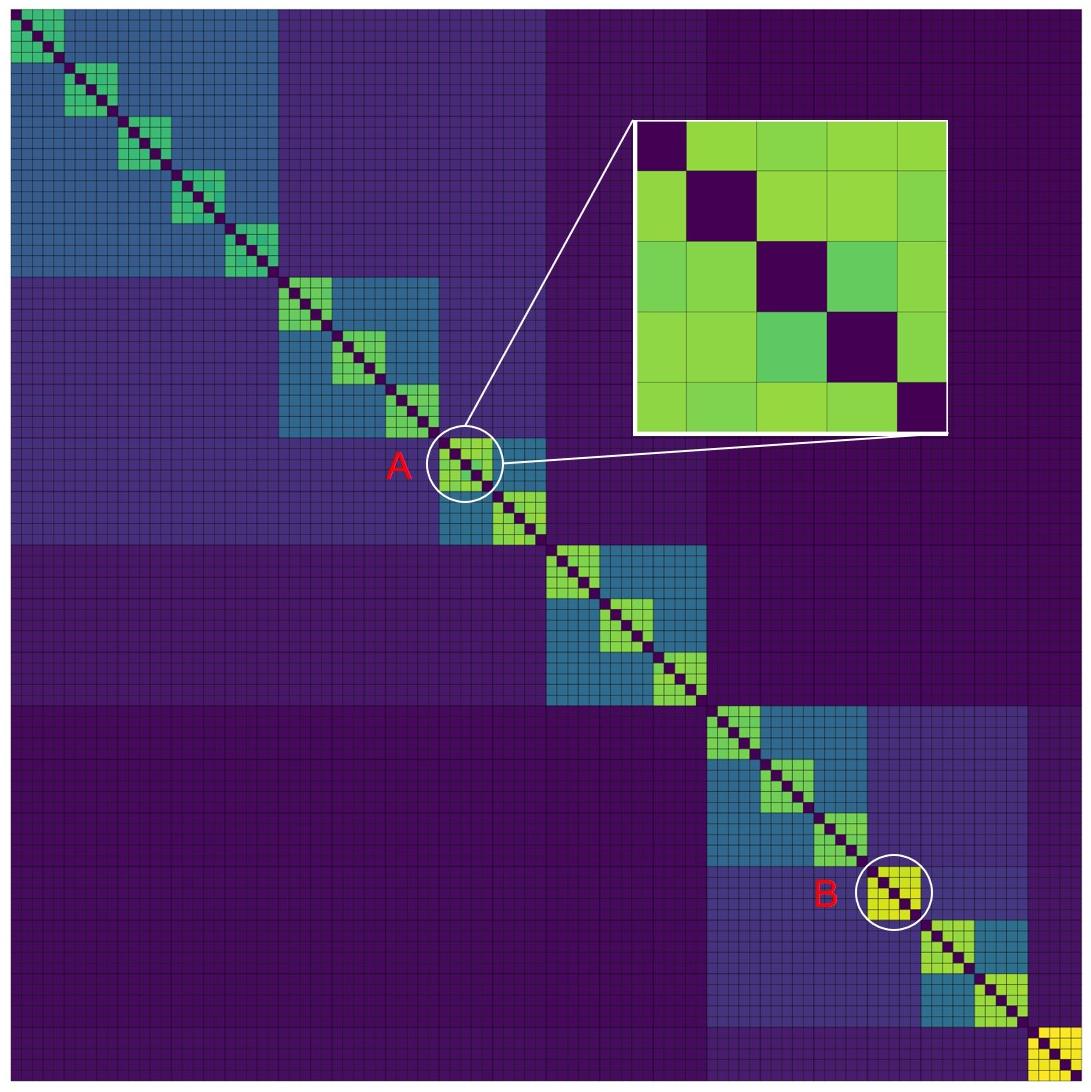}
    \caption{Heat maps of $\mathbf{\Delta}^H$ for CIFAR-100, with colors transitioning from purple to yellow as values increase. This heat map is composed of $100 \times 100$ small squares, which exactly indicates 100 classes (5th level) within CIFAR-100. 20 green or yellow squares, each composed of 5 smaller squares, represent coarser labels for 20 classes (4th level), and larger squares represent coarser classes.} 
    \label{fig:hot_main}
\end{figure}


\subsubsection{Intra-Granularity Difference Visualization.} 
Fig. \ref{fig:hot_main} shows the heat map of the learnable intra-granularity difference matrix $\mathbf{\Delta}^H$, illustrating relationships between the finest-level classes on CIFAR-100. 
Heat map of $\mathbf{\Delta}^H$ for FGVC-Aircraft are provided in the Appendix.
In a specific coarse class (zoomed portion \textbf{A}), the colors of the 3rd and the 4th classes are greener, indicating a weaker correlation compared to other fine-grained classes.
At the coarse-grained level, square (\textbf{B}) is more yellow than square (\textbf{A}), indicating finer-grained classes in (\textbf{B}) share closer relations with each other.
This illustrates that the label tree fails to accurately describe the difference between classes, necessitating the use of $\mathbf{\Delta}^H$ to represent their relationships.

\subsection{Ablation Study}
\subsubsection{Sensitivity analysis of $\alpha$.}
\begin{figure}[!ht]
    \centering
    \subfloat{
        \includegraphics[width=0.45\linewidth]{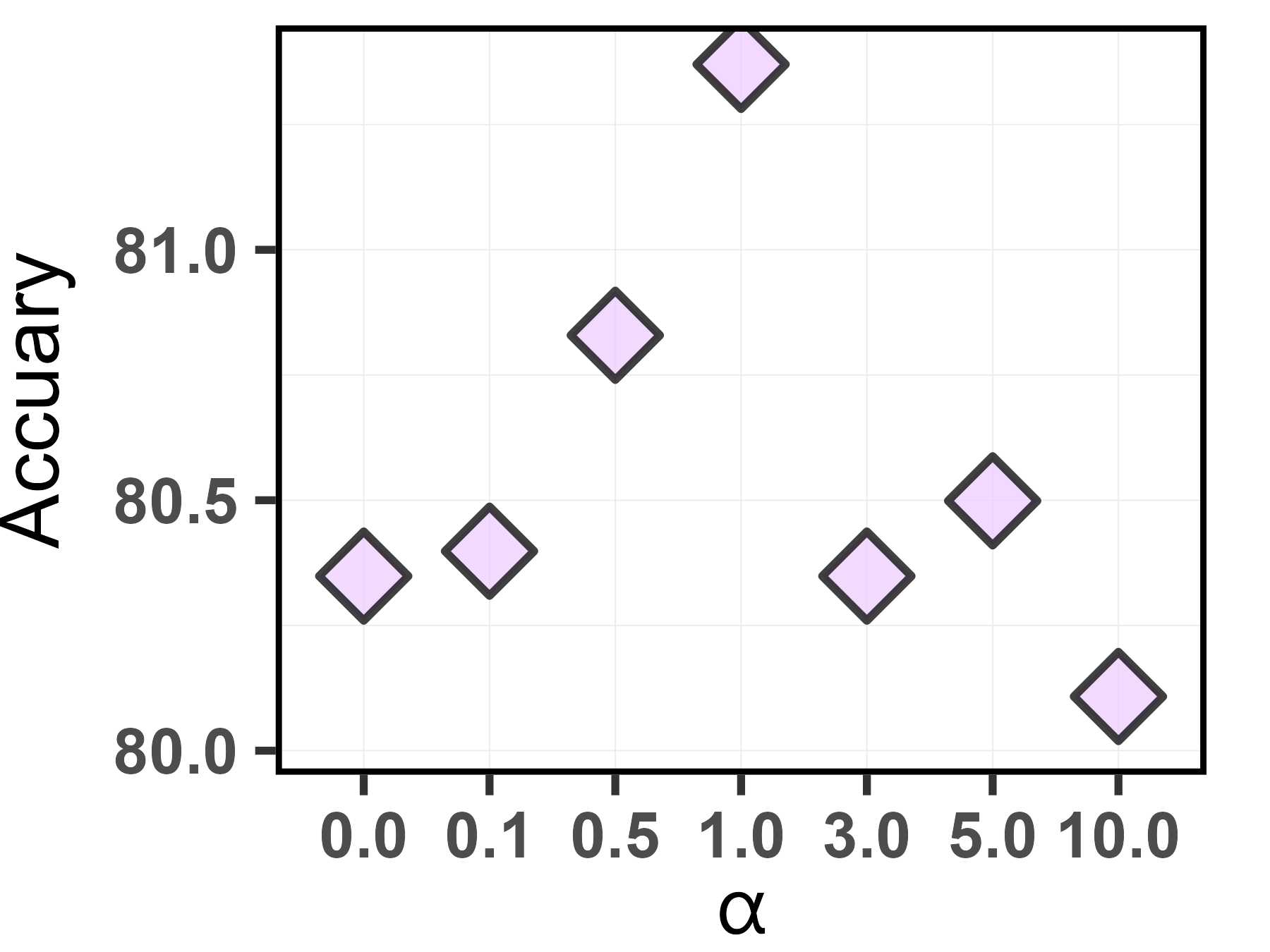}
        \label{fig:fgvc_lambda_acc}
    }
    \subfloat{
        \includegraphics[width=0.45\linewidth]{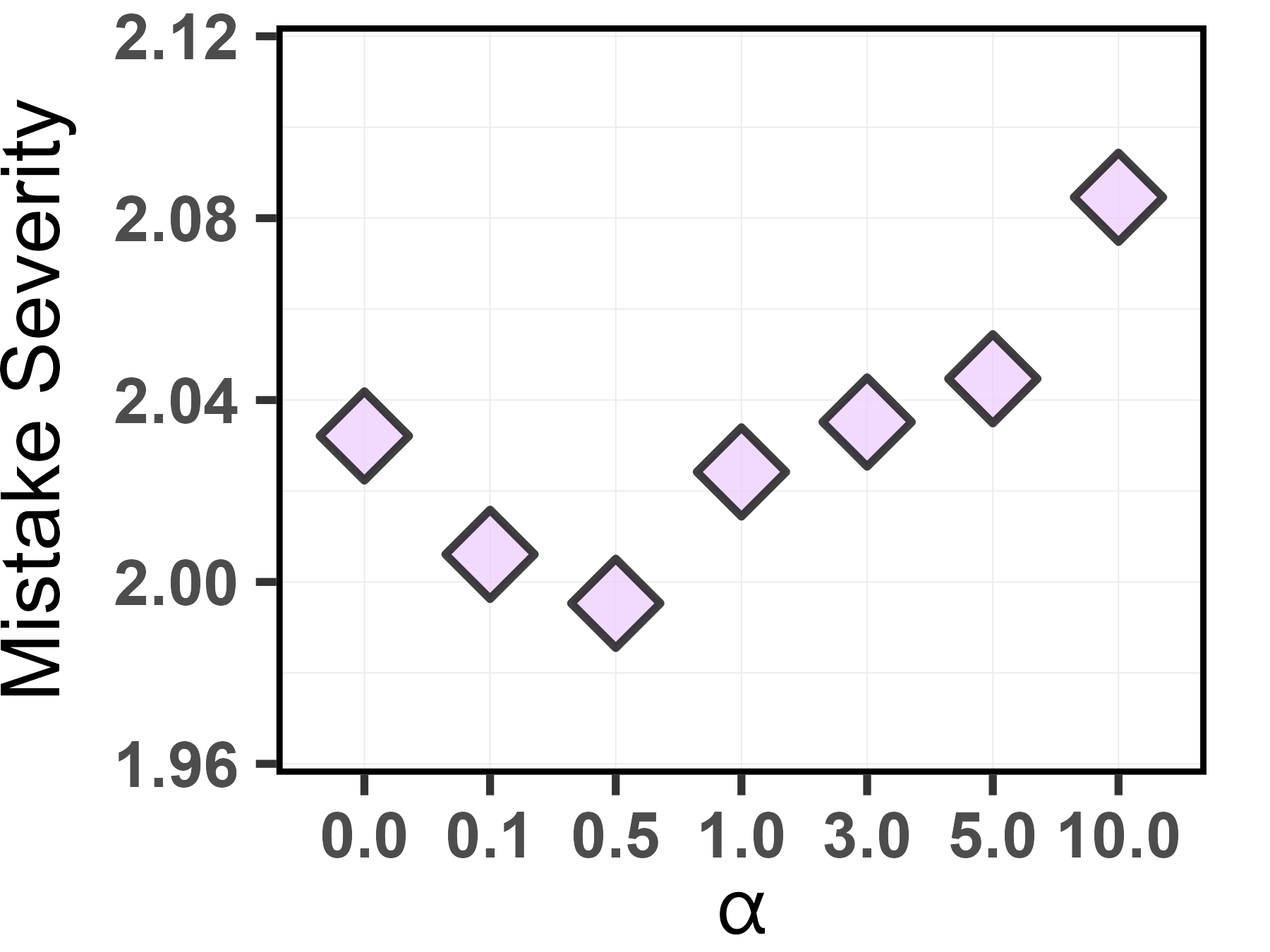}
        \label{fig:fgvc_lambda}
    }\\
    \subfloat{
        \includegraphics[width=0.45\linewidth]{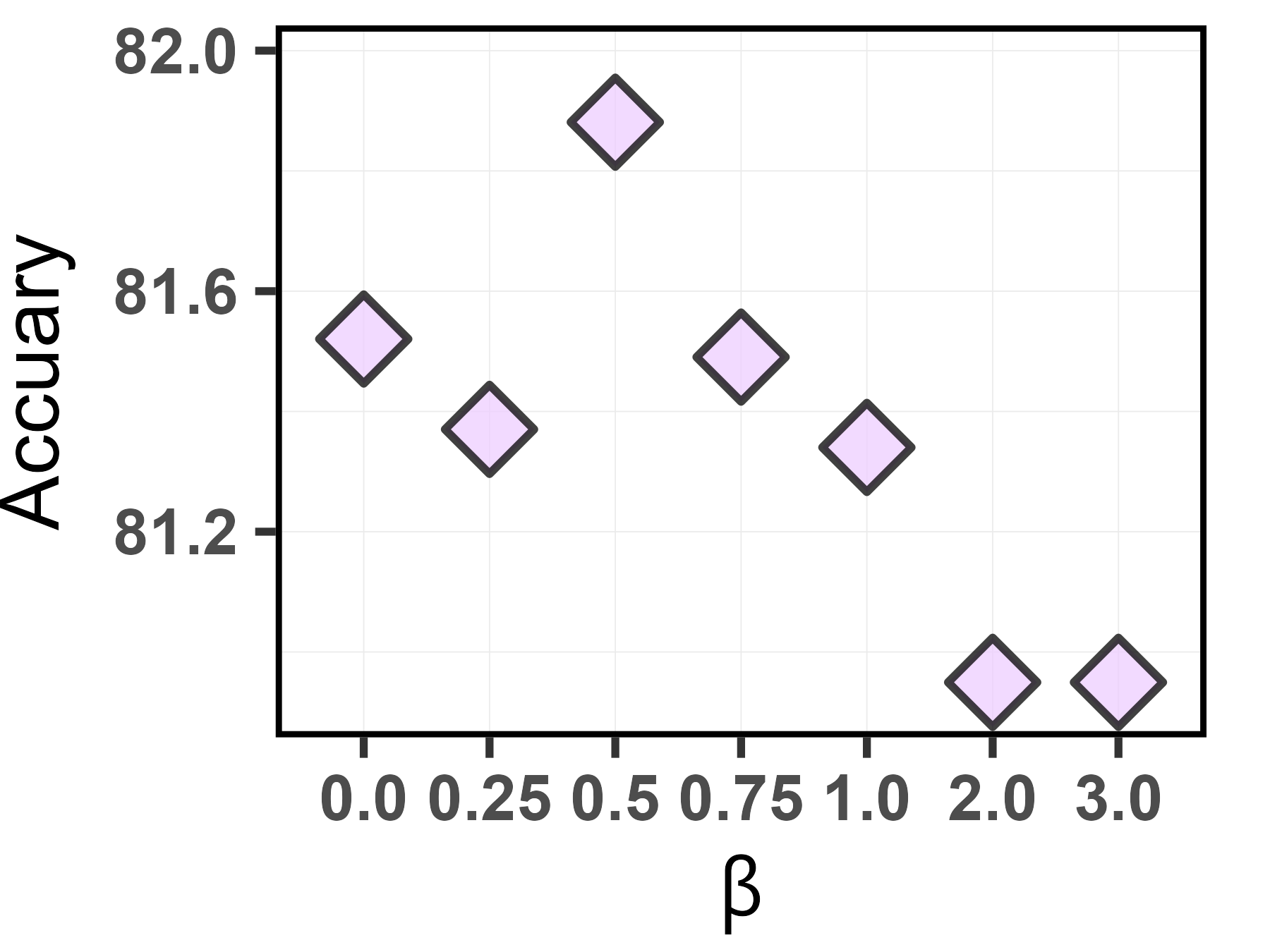}
        \label{fig:fgvc_beta_acc}
    }
    \subfloat{
        \includegraphics[width=0.45\linewidth]{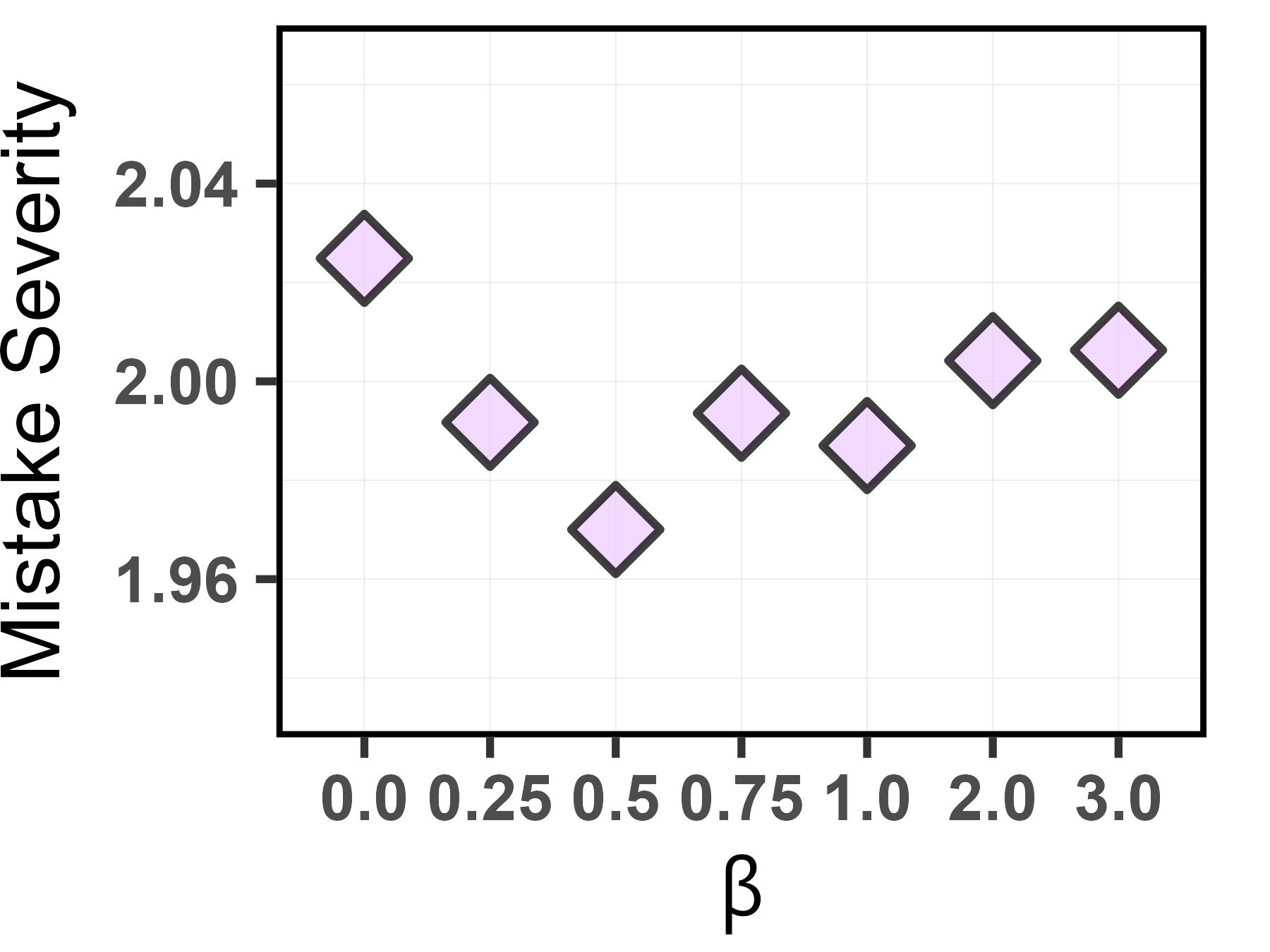}
        \label{fig:fgvc_beta}
    }
    \caption{Sensitivity analysis about $\alpha$ and $\beta$ on FGVC-Aircraft.}
    \label{fig:lambda_main}
\end{figure}

Fig.\ref{fig:lambda_main}  presents the sensitivity analysis of $\alpha$ on the FGVC-Aircraft dataset, while the analysis for the CIFAR-100 dataset is provided in the Appendix.
The weight assigned to level $h$ in the overall optimization objective is given by $\lambda_h = \exp(\alpha \cdot (h-H))$.
A smaller $\alpha$ increases the weights of the coarser levels. 
When $\alpha$ is set to 0, all levels receive equal weight, resulting in a uniform optimization objective.
Within the $\alpha$ range of 0.5 to 1.0, the model reduces Mistake Severity while maintaining Accuracy. 
However, a large $\alpha$ significantly decreases the weight of the coarse level, leading to a notable increase in Mistake Severity.
This phenomenon illustrates the importance of coarse-grained classification for fine-grained learning.

\subsubsection{Sensitivity analysis of $\beta$.}
Fig.\ref{fig:lambda_main}  shows the sensitivity analysis of $\beta$ on the FGVC-Aircraft dataset. 
For the sensitivity analysis of $\beta$ on the CIFAR-100 dataset please see Appendix.
$\beta$ is the weight for the learnable adjustment matrix $\mathbf{\Delta}^h$. 
Too large or too small $\beta$ leads to decreased performance.
When $\beta=0$, the learnable adjustment matrix $\mathbf{\Delta}^h$ has no effect. 
Conversely, too large $\beta$ may cause hierarchical level spanning, where some levels deviate significantly from the originally defined distance matrix after adjustment.
The model achieved the best performance at $\beta=0.5$.


\subsubsection{Sensitivity analysis of $\epsilon$ and $\gamma$.}
\begin{figure}[!ht]
    \centering
    \subfloat[FGVC-Aircraft]{
        \includegraphics[width=0.45\linewidth]{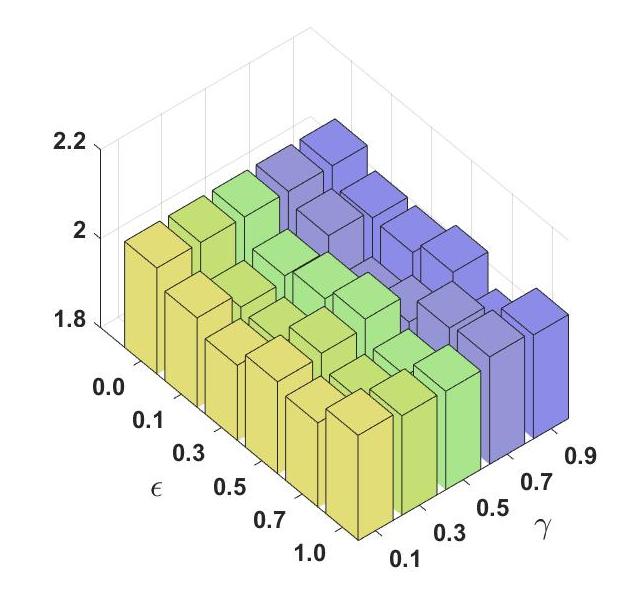}
        \label{fig:fgvc_gam_ep}
    }
    \subfloat[CIFAR-100]{
        \includegraphics[width=0.45\linewidth]{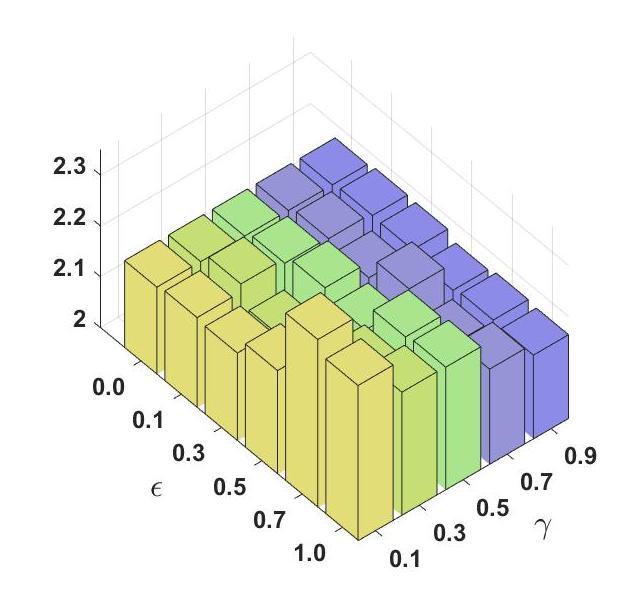}
        \label{fig:cifar100_gam_ep}
    }
    \caption{Sensitivity analysis about $\epsilon$ and $\gamma$ on FGVC-Aircraft and CIFAR-100.}
    \label{fig:gam_ep_main}
\end{figure}

Fig.\ref{fig:gam_ep_main}  shows the sensitivity analysis of $\epsilon$ and $\gamma$ on the FGVC-Aircraft and CIFAR-100 datasets. 
$\gamma$ is the temperature coefficient in the soft label generation process, and $\epsilon$ is the proportion of the soft label. 
When $\epsilon=0$, it means only hard label is used, and when $\epsilon=1$, it means only soft label is used.
$\epsilon$ and $\gamma$ regulate the label encoding together.
In the FGVC-Aircraft dataset, Mistake Severity is low when $\epsilon$ is $0.3$ and $\gamma$ is around $0.7$.
In the CIFAR-100 dataset, Mistake Severity is low when $\epsilon$ and $\gamma$ are both around $0.3$.
It can be observed that using only hard label or only soft label is always not the optimal solution.


\section{Conclusion}
In this paper, we focus on the granularity competition problem between different granularities in fine-grained visual classification (FGVC) tasks.
However, coarse-grained features are naturally easy to learn, leading feature-shared-based methods focus only on coarse features instead of fine-grained ones.
To address this issue, we propose a novel method, the Bidirectional Logits Tree (BiLT), which organizes classifiers from the finest to the coarsest levels, rather than a parallel framework where all classifiers share the same feature vector.
Additionally, we observe that predefined label trees cannot accurately capture semantic differences between labels at the same granularity, and therefore propose Adaptive Intra-Granularity Difference Learning (AIGDL).
This method empowers BiLT to learn the fine-grained semantic differences of classes within the same granularity.
Finally, extensive experiments and visualizations justify the effectiveness of our method.

\section{Acknowledgments}
This work was supported in part by the National Key R\&D Program of China under Grant 2018AAA0102000, in part by National Natural Science Foundation of China: 62236008, U21B2038, U23B2051, 61931008, 62122075, 62206264, and 92370102, in part by Youth Innovation Promotion Association CAS, in part by the Strategic Priority Research Program of the Chinese Academy of Sciences, Grant No. XDB0680000, in part by the Innovation Funding of ICT, CAS under Grant No.E000000, and in part by the Postdoctoral Fellowship Program of CPSF under Grant GZB20240729.

\bibliography{aaai25}

\begin{thebibliography}{50}
\providecommand{\natexlab}[1]{#1}

\bibitem[{Bertinetto et~al.(2020)Bertinetto, Mueller, Tertikas, Samangooei, and Lord}]{bertinetto2020making}
Bertinetto, L.; Mueller, R.; Tertikas, K.; Samangooei, S.; and Lord, N.~A. 2020.
\newblock Making better mistakes: Leveraging class hierarchies with deep networks.
\newblock In \emph{Proceedings of the IEEE/CVF Conference on Computer Vision and Pattern Recognition}, 12506--12515.

\bibitem[{Bukchin et~al.(2021)Bukchin, Schwartz, Saenko, Shahar, Feris, Giryes, and Karlinsky}]{bukchin2021fine}
Bukchin, G.; Schwartz, E.; Saenko, K.; Shahar, O.; Feris, R.; Giryes, R.; and Karlinsky, L. 2021.
\newblock Fine-grained angular contrastive learning with coarse labels.
\newblock In \emph{Proceedings of the IEEE/CVF Conference on Computer Vision and Pattern Recognition}, 8730--8740.

\bibitem[{Chang et~al.(2021)Chang, Pang, Zheng, Ma, Song, and Guo}]{chang2021yourflamingo}
Chang, D.; Pang, K.; Zheng, Y.; Ma, Z.; Song, Y.-Z.; and Guo, J. 2021.
\newblock Your" flamingo" is my" bird": Fine-grained, or not.
\newblock In \emph{Proceedings of the IEEE/CVF Conference on Computer Vision and Pattern Recognition}, 11476--11485.

\bibitem[{Chen et~al.(2022)Chen, Wang, Liu, and Qian}]{chen2022label}
Chen, J.; Wang, P.; Liu, J.; and Qian, Y. 2022.
\newblock Label relation graphs enhanced hierarchical residual network for hierarchical multi-granularity classification.
\newblock In \emph{Proceedings of the IEEE/CVF Conference on Computer Vision and Pattern Recognition}, 4858--4867.

\bibitem[{Collins, Bhatt, and Weller(2022)}]{collins2022eliciting}
Collins, K.~M.; Bhatt, U.; and Weller, A. 2022.
\newblock Eliciting and learning with soft labels from every annotator.
\newblock In \emph{Proceedings of the AAAI Conference on Human Computation and Crowdsourcing}, 40--52.

\bibitem[{Deng et~al.(2009)Deng, Dong, Socher, Li, Li, and Fei-Fei}]{deng2009imagenet}
Deng, J.; Dong, W.; Socher, R.; Li, L.-J.; Li, K.; and Fei-Fei, L. 2009.
\newblock Imagenet: A large-scale hierarchical image database.
\newblock In \emph{IEEE Conference on Computer Vision and Pattern Recognition}, 248--255.

\bibitem[{Du et~al.(2021)Du, Xie, Ma, Chang, Song, and Guo}]{du2021progressive}
Du, R.; Xie, J.; Ma, Z.; Chang, D.; Song, Y.-Z.; and Guo, J. 2021.
\newblock Progressive learning of category-consistent multi-granularity features for fine-grained visual classification.
\newblock \emph{IEEE Transactions on Pattern Analysis and Machine Intelligence}, 44(12): 9521--9535.

\bibitem[{Duda and Hart(1974)}]{Duda1974PatternCA}
Duda, R.~O.; and Hart, P.~E. 1974.
\newblock Pattern classification and scene analysis.
\newblock In \emph{A Wiley-Interscience publication}.

\bibitem[{Frome et~al.(2013)Frome, Corrado, Shlens, Bengio, Dean, Ranzato, and Mikolov}]{frome2013devise}
Frome, A.; Corrado, G.~S.; Shlens, J.; Bengio, S.; Dean, J.; Ranzato, M.; and Mikolov, T. 2013.
\newblock Devise: A deep visual-semantic embedding model.
\newblock In \emph{Advances in Neural Information Processing Systems}.

\bibitem[{Garg, Sani, and Anand(2022)}]{garg2022learningERM}
Garg, A.; Sani, D.; and Anand, S. 2022.
\newblock Learning hierarchy aware features for reducing mistake severity.
\newblock In \emph{European Conference on Computer Vision}, 252--267.

\bibitem[{Gong, Bisht, and Xu(2024)}]{gong2024does}
Gong, X.; Bisht, N.; and Xu, G. 2024.
\newblock Does Label Smoothing Help Deep Partial Label Learning?
\newblock In \emph{International Conference on Machine Learning}.

\bibitem[{Grcic, Gadetsky, and Brbic(2024)}]{grcicfine}
Grcic, M.; Gadetsky, A.; and Brbic, M. 2024.
\newblock Fine-grained Classes and How to Find Them.
\newblock In \emph{International Conference on Machine Learning}, 16275--16294.

\bibitem[{Han et~al.(2024)Han, Xu, Yang, Bao, Wen, Jiang, and Huang}]{han2024aucseg}
Han, B.; Xu, Q.; Yang, Z.; Bao, S.; Wen, P.; Jiang, Y.; and Huang, Q. 2024.
\newblock AUCSeg: AUC-oriented Pixel-level Long-tail Semantic Segmentation.
\newblock In \emph{Advances in Neural Information Processing Systems}.

\bibitem[{He et~al.(2016)He, Zhang, Ren, and Sun}]{he2016deep}
He, K.; Zhang, X.; Ren, S.; and Sun, J. 2016.
\newblock Deep residual learning for image recognition.
\newblock In \emph{Proceedings of the IEEE Conference on Computer Vision and Pattern Recognition}, 770--778.

\bibitem[{Jain, Karthik, and Gandhi(2024)}]{jain2024test}
Jain, K.; Karthik, S.; and Gandhi, V. 2024.
\newblock Test-time amendment with a coarse classifier for fine-grained classification.
\newblock In \emph{Advances in Neural Information Processing Systems}.

\bibitem[{Jiang et~al.(2024)Jiang, Tang, Gao, Du, He, and Li}]{Multimodal_Prompting_FGVC}
Jiang, X.; Tang, H.; Gao, J.; Du, X.; He, S.; and Li, Z. 2024.
\newblock Delving into Multimodal Prompting for Fine-Grained Visual Classification.
\newblock In \emph{Proceedings of the AAAI Conference on Artificial Intelligence}, 2570--2578.

\bibitem[{Karthik et~al.(2021)Karthik, Prabhu, Dokania, and Gandhi}]{karthiknoERM}
Karthik, S.; Prabhu, A.; Dokania, P.~K.; and Gandhi, V. 2021.
\newblock No Cost Likelihood Manipulation at Test Time for Making Better Mistakes in Deep Networks.
\newblock In \emph{International Conference on Learning Representations}.

\bibitem[{Kingma and Ba(2015)}]{AdamICLR}
Kingma, D.~P.; and Ba, J. 2015.
\newblock Adam: {A} Method for Stochastic Optimization.
\newblock In \emph{International Conference on Learning Representations}.

\bibitem[{Krizhevsky, Hinton et~al.(2009)}]{krizhevsky2009learning}
Krizhevsky, A.; Hinton, G.; et~al. 2009.
\newblock Learning multiple layers of features from tiny images.

\bibitem[{Landrieu and Garnot(2021)}]{landrieu2021leveraging}
Landrieu, L.; and Garnot, V. S.~F. 2021.
\newblock Leveraging class hierarchies with metric-guided prototype learning.
\newblock In \emph{British Machine Vision Conference}.

\bibitem[{Liang and Davis(2023)}]{liang2023inducing}
Liang, T.; and Davis, J. 2023.
\newblock Inducing neural collapse to a fixed hierarchy-aware frame for reducing mistake severity.
\newblock In \emph{Proceedings of the IEEE/CVF International Conference on Computer Vision}, 1443--1452.

\bibitem[{Liu et~al.(2024)Liu, Roy, Li, Zhong, Sebe, and Ricci}]{liu2024democratizing}
Liu, M.; Roy, S.; Li, W.; Zhong, Z.; Sebe, N.; and Ricci, E. 2024.
\newblock Democratizing Fine-grained Visual Recognition with Large Language Models.
\newblock In \emph{International Conference on Learning Representations}.

\bibitem[{Liu et~al.(2020)Liu, Chen, Pan, Ngo, Chua, and Jiang}]{liu2020hyperbolic}
Liu, S.; Chen, J.; Pan, L.; Ngo, C.-W.; Chua, T.-S.; and Jiang, Y.-G. 2020.
\newblock Hyperbolic visual embedding learning for zero-shot recognition.
\newblock In \emph{Proceedings of the IEEE/CVF Conference on Computer Vision and Pattern Recognition}, 9273--9281.

\bibitem[{Maji et~al.(2013)Maji, Rahtu, Kannala, Blaschko, and Vedaldi}]{maji2013fine}
Maji, S.; Rahtu, E.; Kannala, J.; Blaschko, M.; and Vedaldi, A. 2013.
\newblock Fine-grained visual classification of aircraft.
\newblock \emph{arXiv preprint arXiv:1306.5151}.

\bibitem[{M{\"u}ller, Kornblith, and Hinton(2019)}]{muller2019does}
M{\"u}ller, R.; Kornblith, S.; and Hinton, G.~E. 2019.
\newblock When does label smoothing help?
\newblock In \emph{Advances in Neural Information Processing Systems}.

\bibitem[{Ni et~al.(2021)Ni, Cheng, Chen, Asakura, Soma, Kato, and Chen}]{ni2021superclass}
Ni, J.; Cheng, W.; Chen, Z.; Asakura, T.; Soma, T.; Kato, S.; and Chen, H. 2021.
\newblock Superclass-conditional gaussian mixture model for learning fine-grained embeddings.
\newblock In \emph{International Conference on Learning Representations}.

\bibitem[{Papyan, Han, and Donoho(2020)}]{papyan2020neuralcollapse}
Papyan, V.; Han, X.; and Donoho, D.~L. 2020.
\newblock Prevalence of neural collapse during the terminal phase of deep learning training.
\newblock \emph{Proceedings of the National Academy of Sciences}, 117(40): 24652--24663.

\bibitem[{Park et~al.(2023)Park, Noh, Oh, Baek, and Ham}]{park2023acls}
Park, H.; Noh, J.; Oh, Y.; Baek, D.; and Ham, B. 2023.
\newblock Acls: Adaptive and conditional label smoothing for network calibration.
\newblock In \emph{Proceedings of the IEEE/CVF International Conference on Computer Vision}, 3936--3945.

\bibitem[{Paszke et~al.(2019)Paszke, Gross, Massa, Lerer, Bradbury, Chanan, Killeen, Lin, Gimelshein, Antiga et~al.}]{paszke2019pytorch}
Paszke, A.; Gross, S.; Massa, F.; Lerer, A.; Bradbury, J.; Chanan, G.; Killeen, T.; Lin, Z.; Gimelshein, N.; Antiga, L.; et~al. 2019.
\newblock Pytorch: An imperative style, high-performance deep learning library.
\newblock In \emph{Advances in Neural Information Processing Systems}.

\bibitem[{Pu et~al.(2024)Pu, Han, Wang, Feng, Deng, and Huang}]{tip/PuHWFDH24}
Pu, Y.; Han, Y.; Wang, Y.; Feng, J.; Deng, C.; and Huang, G. 2024.
\newblock Fine-grained recognition with learnable semantic data augmentation.
\newblock \emph{IEEE Transactions on Image Processing}, 33: 3130--3144.

\bibitem[{Qin et~al.(2021)Qin, Wang, Beutel, and Chi}]{qin2021improving}
Qin, Y.; Wang, X.; Beutel, A.; and Chi, E. 2021.
\newblock Improving calibration through the relationship with adversarial robustness.
\newblock In \emph{Advances in Neural Information Processing Systems}, 14358--14369.

\bibitem[{Ren et~al.(2018)Ren, Triantafillou, Ravi, Snell, Swersky, Tenenbaum, Larochelle, and Zemel}]{ren2018meta}
Ren, M.; Triantafillou, E.; Ravi, S.; Snell, J.; Swersky, K.; Tenenbaum, J.~B.; Larochelle, H.; and Zemel, R.~S. 2018.
\newblock Meta-learning for semi-supervised few-shot classification.
\newblock In \emph{International Conference on Learning Representations}.

\bibitem[{Silla and Freitas(2011)}]{silla2011survey}
Silla, C.~N.; and Freitas, A.~A. 2011.
\newblock A survey of hierarchical classification across different application domains.
\newblock \emph{Data Mining and Knowledge Discovery}, 22: 31--72.

\bibitem[{Touvron et~al.(2021)Touvron, Sablayrolles, Douze, Cord, and J{\'e}gou}]{touvron2021grafit}
Touvron, H.; Sablayrolles, A.; Douze, M.; Cord, M.; and J{\'e}gou, H. 2021.
\newblock Grafit: Learning fine-grained image representations with coarse labels.
\newblock In \emph{Proceedings of the IEEE/CVF International Conference on Computer Vision}, 874--884.

\bibitem[{Van~Horn et~al.(2018)Van~Horn, Mac~Aodha, Song, Cui, Sun, Shepard, Adam, Perona, and Belongie}]{van2018inaturalist}
Van~Horn, G.; Mac~Aodha, O.; Song, Y.; Cui, Y.; Sun, C.; Shepard, A.; Adam, H.; Perona, P.; and Belongie, S. 2018.
\newblock The inaturalist species classification and detection dataset.
\newblock In \emph{Proceedings of the IEEE/CVF International Conference on Computer Vision}, 8769--8778.

\bibitem[{Wang et~al.(2015)Wang, Shen, Shao, Zhang, Xue, and Zhang}]{wang2015multiple}
Wang, D.; Shen, Z.; Shao, J.; Zhang, W.; Xue, X.; and Zhang, Z. 2015.
\newblock Multiple granularity descriptors for fine-grained categorization.
\newblock In \emph{Proceedings of the IEEE International Conference on Computer Vision}, 2399--2406.

\bibitem[{Wang et~al.(2023)Wang, Zou, Zhang, Zhu, and Jing}]{mm/WangZZZJ23}
Wang, R.; Zou, C.; Zhang, W.; Zhu, Z.; and Jing, L. 2023.
\newblock Consistency-aware Feature Learning for Hierarchical Fine-grained Visual Classification.
\newblock In \emph{Proceedings of the 31st {ACM} International Conference on Multimedia}, 2326--2334.

\bibitem[{Wang et~al.(2024)Wang, Sun, Li, and Yang}]{wang2024transhp}
Wang, W.; Sun, Y.; Li, W.; and Yang, Y. 2024.
\newblock Transhp: Image classification with hierarchical prompting.
\newblock In \emph{Advances in Neural Information Processing Systems}.

\bibitem[{Wang et~al.(2021)Wang, Wang, Hu, Zhou, and Su}]{wang2021hierarchical}
Wang, Y.; Wang, Z.; Hu, Q.; Zhou, Y.; and Su, H. 2021.
\newblock Hierarchical semantic risk minimization for large-scale classification.
\newblock \emph{IEEE Transactions on Cybernetics}, 52(9): 9546--9558.

\bibitem[{Wei et~al.(2022)Wei, Liu, Liu, Niu, Sugiyama, and Liu}]{wei2022smooth}
Wei, J.; Liu, H.; Liu, T.; Niu, G.; Sugiyama, M.; and Liu, Y. 2022.
\newblock To Smooth or Not? When Label Smoothing Meets Noisy Labels.
\newblock In \emph{International Conference on Machine Learning}, 23589--23614.

\bibitem[{Wu et~al.(2016)Wu, Merler, Uceda-Sosa, and Smith}]{wu2016learning}
Wu, H.; Merler, M.; Uceda-Sosa, R.; and Smith, J.~R. 2016.
\newblock Learning to make better mistakes: Semantics-aware visual food recognition.
\newblock In \emph{Proceedings of the 24th ACM International Conference on Multimedia}, 172--176.

\bibitem[{Xian et~al.(2016)Xian, Akata, Sharma, Nguyen, Hein, and Schiele}]{xian2016latent}
Xian, Y.; Akata, Z.; Sharma, G.; Nguyen, Q.; Hein, M.; and Schiele, B. 2016.
\newblock Latent embeddings for zero-shot classification.
\newblock In \emph{Proceedings of the IEEE Conference on Computer Vision and Pattern Recognition}, 69--77.

\bibitem[{Xu et~al.(2023)Xu, Sun, Zhang, Xu, Wei, and Yang}]{HyperbolicLearningfromcoarse}
Xu, S.; Sun, Y.; Zhang, F.; Xu, A.; Wei, X.; and Yang, Y. 2023.
\newblock Hyperbolic Space with Hierarchical Margin Boosts Fine-Grained Learning from Coarse Labels.
\newblock In \emph{Advances in Neural Information Processing Systems}.

\bibitem[{Yuan et~al.(2020)Yuan, Tay, Li, Wang, and Feng}]{yuan2020revisiting}
Yuan, L.; Tay, F.~E.; Li, G.; Wang, T.; and Feng, J. 2020.
\newblock Revisiting knowledge distillation via label smoothing regularization.
\newblock In \emph{Proceedings of the IEEE/CVF Conference on Computer Vision and Pattern Recognition}, 3903--3911.

\bibitem[{Zagoruyko and Komodakis(2016)}]{zagoruyko2016wide}
Zagoruyko, S.; and Komodakis, N. 2016.
\newblock Wide Residual Networks.
\newblock In \emph{British Machine Vision Conference}.

\bibitem[{Zhang et~al.(2021)Zhang, Jiang, Hou, Wei, Han, Li, and Cheng}]{zhang2021delving}
Zhang, C.-B.; Jiang, P.-T.; Hou, Q.; Wei, Y.; Han, Q.; Li, Z.; and Cheng, M.-M. 2021.
\newblock Delving deep into label smoothing.
\newblock \emph{IEEE Transactions on Image Processing}, 30: 5984--5996.

\bibitem[{Zhang et~al.(2023)Zhang, Chen, Zhang, Xiao, and Li}]{zhang2023diffsmooth}
Zhang, J.; Chen, Z.; Zhang, H.; Xiao, C.; and Li, B. 2023.
\newblock $\{$DiffSmooth$\}$: Certifiably robust learning via diffusion models and local smoothing.
\newblock In \emph{32nd USENIX Security Symposium (USENIX Security 23)}, 4787--4804.

\bibitem[{Zhang and Zhou(2013)}]{zhang2013review}
Zhang, M.-L.; and Zhou, Z.-H. 2013.
\newblock A review on multi-label learning algorithms.
\newblock \emph{IEEE Transactions on Knowledge and Data Engineering}, 26(8): 1819--1837.

\bibitem[{Zhang et~al.(2024)Zhang, Zheng, Shui, and Yang}]{zhang2024hls}
Zhang, S.; Zheng, S.; Shui, Z.; and Yang, L. 2024.
\newblock HLS-FGVC: Hierarchical Label Semantics Enhanced Fine-Grained Visual Classification.
\newblock In \emph{IEEE International Conference on Acoustics, Speech and Signal Processing}, 7370--7374.

\bibitem[{Zhao, Li, and Xing(2011)}]{zhao2011large}
Zhao, B.; Li, F.; and Xing, E. 2011.
\newblock Large-scale category structure aware image categorization.
\newblock In \emph{Advances in Neural Information Processing Systems}.

\end{thebibliography}

\twocolumn
\appendix
\section{Appendix}
\subsection{Related Work}
\subsubsection{Fine-grained Category Discovery Methods}
These methods primarily focus on discovering fine-grained categories using coarse-grained information.
\cite{bukchin2021fine} was the first to address the problem of identifying fine-grained categories from coarse-grained information. 
They proposed a method combining coarse-grained pretraining and self-supervised contrastive learning, marking the first attempt to integrate these approaches for this task.
Grafit \cite{touvron2021grafit} later combined coarse-grained labels with fine-grained latent spaces to enhance fine-grained retrieval and classification accuracy. 
SCGM \cite{ni2021superclass} used the Gaussian Mixture Models to link coarse-grained and fine-grained categories.
More recently, FALCON \cite{grcicfine} addressed a more challenging scenario, discovering latent fine-grained categories without prior observations.
They tackled this problem using an alternating optimization approach and gave a rich theoretical analysis. 
Since our method relies on multi-granularity category, these approaches provide a foundation for our method.

\subsection{Experiments}
\subsubsection{Datasets}
For FGVC-Aircraft, we use the original hierarchical labels and the split setting provided by the dataset. 
For CIFAR-100, we use the hierarchical labels from \cite{landrieu2021leveraging} and the split setting from \cite{garg2022learningERM}. 
For iNaturalist2019, we use the hierarchical labels and split setting provided by \cite{bertinetto2020making}. 
For tieredImageNet-H, we use the hierarchical labels and split setting provided by \cite{ren2018meta}. 
For all datasets, the distance between any two nodes in the label tree is defined using the Lowest Common Ancestor (LCA). 
Dataset statistics are shown in Tab. \ref{tab:dataset}.
\begin{table}[!h]   
    \centering
    \setlength{\tabcolsep}{1mm}{
        \begin{tabular}{c|ccccc}
            \toprule
            Datasets         & Levels & Classes & Train   & Val    & Test \\
            \midrule
            FGVC-Aircraft   & 3     & 100     & 3,334   & 3,333  & 3,333 \\
            CIFAR-100       & 5     & 100     & 45,000  & 5,000  & 10,000 \\
            iNaturalist2019 & 7     & 1010    & 187,385 & 40,121 & 40,737 \\
            tieredImageNet-H & 12   & 608     & 425,600 & 15,200 & 15,200 \\
            \bottomrule
        \end{tabular}%
    }
    \caption{Statistics of the datasets.}
    \label{tab:dataset}
\end{table}
\begin{table*}[!t]
    \centering
    \resizebox{0.95\textwidth}{!}{
    \begin{tabular}{c|c||ccc||c}
        \toprule
        Method & Mistake Severity($\downarrow$) & Hier Dist@1($\downarrow$) & Hier Dist@5($\downarrow$) & Hier Dist@20($\downarrow$) & Top-1 Accuracy($\uparrow$)\\
        \midrule
        Cross-Entropy & \cellcolor[rgb]{ .999,  .999,  .999}{2.29 +/- 0.0185}           & \cellcolor[rgb]{ .988,  .969,  .969}{0.67 +/- 0.0080}           & \cellcolor[rgb]{ .999,  .999,  .999}{1.98 +/- 0.0029}           & \cellcolor[rgb]{ .999,  .999,  .999}{3.41 +/- 0.0070}           & \cellcolor[rgb]{ .988,  .949,  .953}{70.66 +/- 0.2274} \\
        HXE           & \cellcolor[rgb]{ .999,  .999,  .999}{2.29 +/- 0.0206}           & \cellcolor[rgb]{ .999,  .999,  .999}{0.75 +/- 0.0121}           & \cellcolor[rgb]{ .996,  .988,  .988}{1.84 +/- 0.0082}           & \cellcolor[rgb]{ .988,  .965,  .965}{2.41 +/- 0.0039}           & \cellcolor[rgb]{ .999,  .999,  .999}{67.16 +/- 0.3120}\\
        Soft-Labels   & \cellcolor[rgb]{ .992,  .973,  .973}{2.19 +/- 0.0133}           & \cellcolor[rgb]{ .992,  .980,  .980}{0.71 +/- 0.0099}           & \cellcolor[rgb]{ .988,  .957,  .957}{1.28 +/- 0.0071}           & \cellcolor[rgb]{ .980,  .941,  .945}{2.04 +/- 0.0085}           & \cellcolor[rgb]{ .996,  .98,  .984}{68.47 +/- 0.2941}\\
        Flamingo      & \cellcolor[rgb]{ .988,  .957,  .957}{2.13 +/- 0.0063}           & \cellcolor[rgb]{ .988,  .957,  .957}{0.64 +/- 0.0014}           & \cellcolor[rgb]{ .996,  .988,  .988}{1.79 +/- 0.0126}           & \cellcolor[rgb]{ .996,  .992,  .992}{3.28 +/- 0.0110}           & \cellcolor[rgb]{ .988,  .949,  .953}{70.67 +/- 0.2095}\\
        CRM           & \cellcolor[rgb]{ .996,  .984,  .984}{2.24 +/- 0.0155}           & \cellcolor[rgb]{ .988,  .965,  .965}{0.66 +/- 0.0062}           & \cellcolor[rgb]{ .969,  .902,  .902}{1.19 +/- 0.0046}           & \cellcolor[rgb]{ .957,  .875,  .878}{1.76 +/- 0.0050}           & \cellcolor[rgb]{ .988,  .949,  .953}{70.66 +/- 0.2274}   \\
        HAF           & \cellcolor[rgb]{ .988,  .957,  .957}{2.13 +/- 0.0192}           & \cellcolor[rgb]{ .980,  .941,  .941}{0.63 +/- 0.0045}           & \cellcolor[rgb]{ .988,  .969,  .969}{1.55 +/- 0.2188}           & \cellcolor[rgb]{ .992,  .973,  .973}{2.68 +/- 0.4208}           & \cellcolor[rgb]{ .988,  .949,  .953}{70.57 +/- 0.1645}\\
        HAFrame       & \cellcolor[rgb]{ .961,  .886,  .886}{2.06 +/- 0.0087}           & \cellcolor[rgb]{ .965,  .894,  .898}{0.60 +/- 0.0030} & \cellcolor[rgb]{ .957,  .871,  .875}\underline{1.14 +/- 0.0025} & \cellcolor[rgb]{ .957,  .871,  .875}\underline{1.74 +/- 0.0017} & \cellcolor[rgb]{ .976,  .914,  .918}{70.89 +/- 0.1759} \\
        HiE           & \cellcolor[rgb]{ .957,  .867,  .871}\underline{2.04 +/- 0.0162} & \cellcolor[rgb]{ .957,  .867,  .871}\textbf{0.58 +/- 0.0041}    & \cellcolor[rgb]{ .965,  .894,  .898}{1.18 +/- 0.1247}           & \cellcolor[rgb]{ .988,  .957,  .957}{2.09 +/- 0.2103}           & \cellcolor[rgb]{ .933,  .792,  .796}\textbf{71.43 +/- 0.2584}\\
        Ours          & \cellcolor[rgb]{ .957,  .867,  .871}\textbf{2.04 +/- 0.0070}    & \cellcolor[rgb]{ .965,  .894,  .898}\underline{0.59 +/- 0.0019} & \cellcolor[rgb]{ .957,  .867,  .871}\textbf{1.13 +/- 0.0017}    & \cellcolor[rgb]{ .957,  .867,  .871}\textbf{1.72 +/- 0.0018}    & \cellcolor[rgb]{ .984,  .945,  .949}\underline{70.98 +/- 0.0450} \\
        \bottomrule
    \end{tabular}%
    }
    \caption{Performance comparisons on the iNaturalist2019 dataset with different metrics. The first and second best results are highlighted with \textbf{bold} text and \underline{underline}, respectively.}
    \label{tab:iNat19}
\end{table*}%

\begin{table*}[!t]
    \centering
    \resizebox{0.95\textwidth}{!}{
    \begin{tabular}{c|c||ccc||c}
        \toprule
        Method & Mistake Severity($\downarrow$) & Hier Dist@1($\downarrow$) & Hier Dist@5($\downarrow$) & Hier Dist@20($\downarrow$) & Top-1 Accuracy($\uparrow$)\\
        \midrule
        Cross-Entropy & \cellcolor[rgb]{ .999,  .999,  .999}{6.95 +/- 0.0208} & \cellcolor[rgb]{ .808,  .808,  .961}{1.83 +/- 0.0117} & \cellcolor[rgb]{ .941,  .941,  .988}{5.69 +/- 0.0192}& \cellcolor[rgb]{ .969,  .969,  .992}{7.34 +/- 0.0291} & \cellcolor[rgb]{ .776,  .776,  .957}{73.63 +/- 0.1165} \\
        HXE           & \cellcolor[rgb]{ .925,  .925,  .984}{6.93 +/- 0.0297} & \cellcolor[rgb]{ .773,  .773,  .953}{1.81 +/- 0.0109} & \cellcolor[rgb]{ .961,  .961,  .992}{5.71 +/- 0.0101}& \cellcolor[rgb]{ .816,  .816,  .961}{6.99 +/- 0.0091} & \cellcolor[rgb]{ .933,  .933,  .988}{71.29 +/- 0.2378} \\
        Soft-Labels   & \cellcolor[rgb]{ .965,  .965,  .992}{6.94 +/- 0.0263} & \cellcolor[rgb]{ .788,  .788,  .957}{1.82 +/- 0.0117} & \cellcolor[rgb]{ .925,  .925,  .984}{5.67 +/- 0.0099}& \cellcolor[rgb]{ .788,  .788,  .957}{6.92 +/- 0.0109} & \cellcolor[rgb]{ .999,  .999,  .999}{70.18 +/- 0.2063} \\
        Flamingo      & \cellcolor[rgb]{ .925,  .925,  .984}{6.93 +/- 0.0391} & \cellcolor[rgb]{ .999,  .999,  .999}{1.92 +/- 0.0135} & \cellcolor[rgb]{ .999,  .999,  .999}{5.75 +/- 0.0130}& \cellcolor[rgb]{ .999,  .999,  .999}{7.41 +/- 0.0098} & \cellcolor[rgb]{ .867,  .867,  .973}{72.34 +/- 0.1488} \\
        CRM           & \cellcolor[rgb]{ .788,  .788,  .957}{6.89 +/- 0.0272} & \cellcolor[rgb]{ .788,  .788,  .957}{1.82 +/- 0.0155} & \cellcolor[rgb]{ .675,  .675,  .937}\textbf{4.82 +/- 0.0062}& \cellcolor[rgb]{ .675,  .675,  .937}\textbf{6.03 +/- 0.0041} & \cellcolor[rgb]{ .788,  .788,  .957}{73.54 +/- 0.1495} \\
        HAF           & \cellcolor[rgb]{ .788,  .788,  .957}{6.89 +/- 0.0281} & \cellcolor[rgb]{ .788,  .788,  .957}{1.82 +/- 0.0125} & \cellcolor[rgb]{ .788,  .788,  .957}{5.52 +/- 0.0176}& \cellcolor[rgb]{ .800,  .800,  .957}{6.95 +/- 0.0120} & \cellcolor[rgb]{ .788,  .788,  .957}{73.52 +/- 0.1613} \\
        HAFrame       & \cellcolor[rgb]{ .788,  .788,  .957}{6.89 +/- 0.0251} & \cellcolor[rgb]{ .745,  .745,  .949}\underline{1.79 +/- 0.0216} & \cellcolor[rgb]{ .690,  .690,  .937}\underline{4.94 +/- 0.0118}& \cellcolor[rgb]{ .686,  .686,  .937}\underline{6.15 +/- 0.0065} & \cellcolor[rgb]{ .729,  .729,  .949}\underline{74.00 +/- 0.3549} \\
        HiE           & \cellcolor[rgb]{ .741,  .741,  .945}\underline{6.85 +/- 0.0306} & \cellcolor[rgb]{ .827,  .827,  .965}{1.84 +/- 0.0189} & \cellcolor[rgb]{ .741,  .741,  .949}{5.25 +/- 0.0143}& \cellcolor[rgb]{ .765,  .765,  .949}{6.74 +/- 0.0099} & \cellcolor[rgb]{ .835,  .835,  .969}{72.78 +/- 0.2512} \\
        Ours          & \cellcolor[rgb]{ .675,  .675,  .937}\textbf{6.77 +/- 0.0371} & \cellcolor[rgb]{ .675,  .675,  .937}\textbf{1.72 +/- 0.0241} & \cellcolor[rgb]{ .769,  .769,  .953}{5.41 +/- 0.0232} & \cellcolor[rgb]{ .78,  .78,  .953}{6.86 +/- 0.0158} & \cellcolor[rgb]{ .675,  .675,  .937}\textbf{74.41 +/- 0.3347} \\
        \bottomrule
    \end{tabular}
    }
    \caption{Performance comparisons on the tieredImageNet-H dataset with different metrics. The first and second best results are highlighted with \textbf{bold} text and \underline{underline}, respectively.}
    \label{tab:imagenet}
\end{table*}

\subsubsection{Evaluation Metrics.}
We evaluate the performance of the methods by the following metrics:
\begin{itemize}
    \item \textbf{Mistake Severity} is proposed in \cite{bertinetto2020making}. 
    \begin{equation*}
        \mathrm{\textbf{Mistake  Severity}} = \frac{\mathrm{LCA}( \widehat{\mathcal{Y}}^H , \mathcal{Y}^H )}{ |\mathcal{Y}^H| - |\widehat{\mathcal{Y}}^H \cap \mathcal{Y}^H| }, 
    \end{equation*}
    where $\widehat{\mathcal{Y}}^H$ is the predicted label set, $\mathcal{Y}^H$ is the ground truth label set, and $|\cdot|$ means the size of set.
    It measures the average LCA distance between the incorrectly predicted label and the ground truth label in the label tree, which reflects the severity of the mistakes made by the model.
    
    \item \textbf{Hier Dist@$k$}, also proposed in \cite{bertinetto2020making}, measures the average LCA distance between the top-$k$ predicted labels and the ground truth label in the label tree, respectively.
    Given the top-$k$ predicted label set $\widehat{\mathcal{Y}}^H_k$ and the top-$k$ ground truth label set $\mathcal{Y}^H_k$, the definition of \textbf{Hier Dist@$k$} is 
    \begin{equation*}
        \mathrm{\textbf{Hier Dist}}@k = \frac{\mathrm{LCA}( \widehat{\mathcal{Y}}^H_k , \mathcal{Y}^H_k )}{ |\mathcal{Y}^H|}, 
    \end{equation*}
    It reflects the overall quality of the top-$k$ predictions, which is important for certain downstream tasks.
    
    \item \textbf{Top-1 Accuracy}, a commonly used metric in fine-grained visual classification tasks, defined as 
    \begin{equation*}
        \mathrm{\textbf{Top-1 Accuracy}} = \frac{|\widehat{\mathcal{Y}}^H \cap \mathcal{Y}^H|}{ |\mathcal{Y}^H|}.
    \end{equation*}
\end{itemize}

\begin{figure}[!t]
    \centering
    \includegraphics[width=0.75\columnwidth]{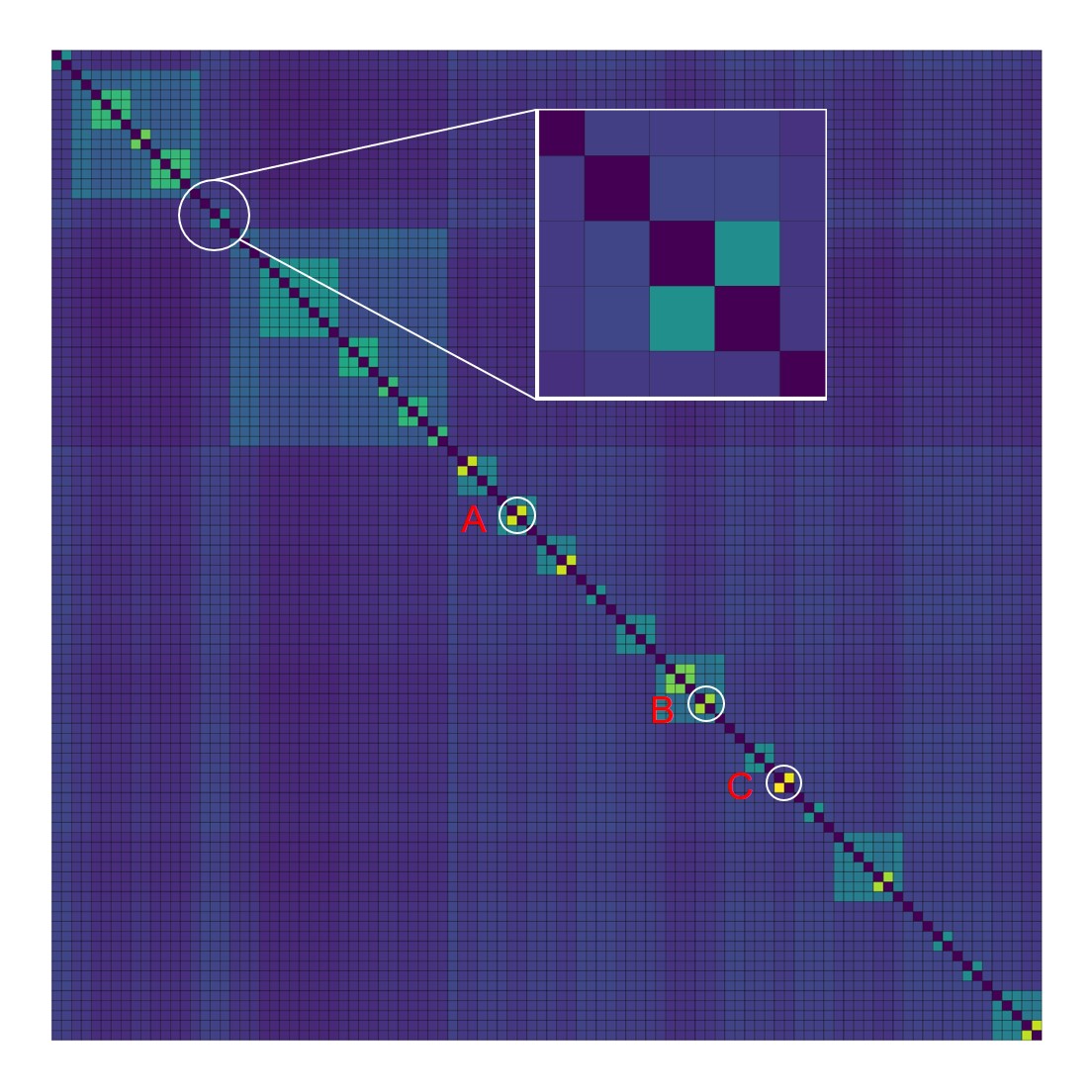}
    \caption{Heat maps of $\mathbf{\Delta}^H$ for FGVC-Aircraft, with colors transitioning from purple to yellow as values increase. } 
    \label{fig:hot_main_fgvc}
\end{figure}

\subsubsection{Implementation Details.}
We implemented our model with PyTorch\footnote{https://pytorch.org/} \cite{paszke2019pytorch} and all experiments were conducted on NVIDIA Tesla A100 80G GPUs.
For CIFAR-100, we use WideResNet-28 \cite{zagoruyko2016wide} as the backbone network for all methods. 
For FGVC-Aircraft and iNaturalist2019, we use ResNet-50 \cite{he2016deep} as the backbone network for all methods and initialize it with the pre-trained model from ImageNet \cite{deng2009imagenet}. 
Models are trained for 200 epochs on CIFAR-100 and 100 epochs on FGVC-Aircraft and iNaturalist2019.
All methods, except for HXE and Soft-Label, use the SGD optimizer with a momentum of 0.9 and a weight decay of 0.0005. 
For HXE and Soft-Label, following \cite{garg2022learningERM}, the Adam \cite{AdamICLR} optimizer is used with a learning rate of 0.001. 
For SGD training, the learning rate is initialized to 0.01 for the backbone network and 0.1 for the transformation layer and classifier. 
All methods are trained with a cosine learning rate scheduler as in \cite{chang2021yourflamingo}. 
We use a batch size of 64 for CIFAR-100 and FGVC-Aircraft, and 256 for iNaturalist2019. 
The same data augmentation strategy as in \cite{garg2022learningERM} is used for all datasets. 
We select the best model based on the validation set and report the results on the test set. 
We run each method 5 times with different random seeds (0-4), and results are presented with a $95\%$ confidence interval following \cite{liang2023inducing}. 

\subsection{Performance Comparisons}
\subsubsection{Intra-Granularity Difference Visualization.}
Fig. \ref{fig:hot_main_fgvc} shows heat maps of the learnable intra-granularity difference matrix $\mathbf{\Delta}^H$ for the finest-level classes on FGVC-Aircraft. 
Same as CIFAR-100, in a specific coarse class (the zoomed portion in Figure), the colors of the 2nd and the 3rd classes are bluer, indicating a weaker correlation compared to other fine-grained classes.
At the coarse-grained level, square (\textbf{B}) is greener than square (\textbf{A} and \textbf{C}), indicating finer-grained classes in (\textbf{A} and \textbf{C}) share closer relations with each other.

\subsection{Ablation Study}
\subsubsection{Sensitivity analysis of $\alpha$.}
Fig.\ref{fig:cifar100_lambda_main}  presents the sensitivity analysis of $\alpha$ on the CIFAR-100 dataset. 
Similar to the FGVC-Aircraft dataset, setting $\alpha$ within $0.5$ to $3.0$ can enhance fine-grained learning by leveraging coarse-grained classification.

\subsubsection{Sensitivity analysis of $\beta$.}
\begin{figure}[!ht]
    \centering
    \subfloat{
        \includegraphics[width=0.45\linewidth]{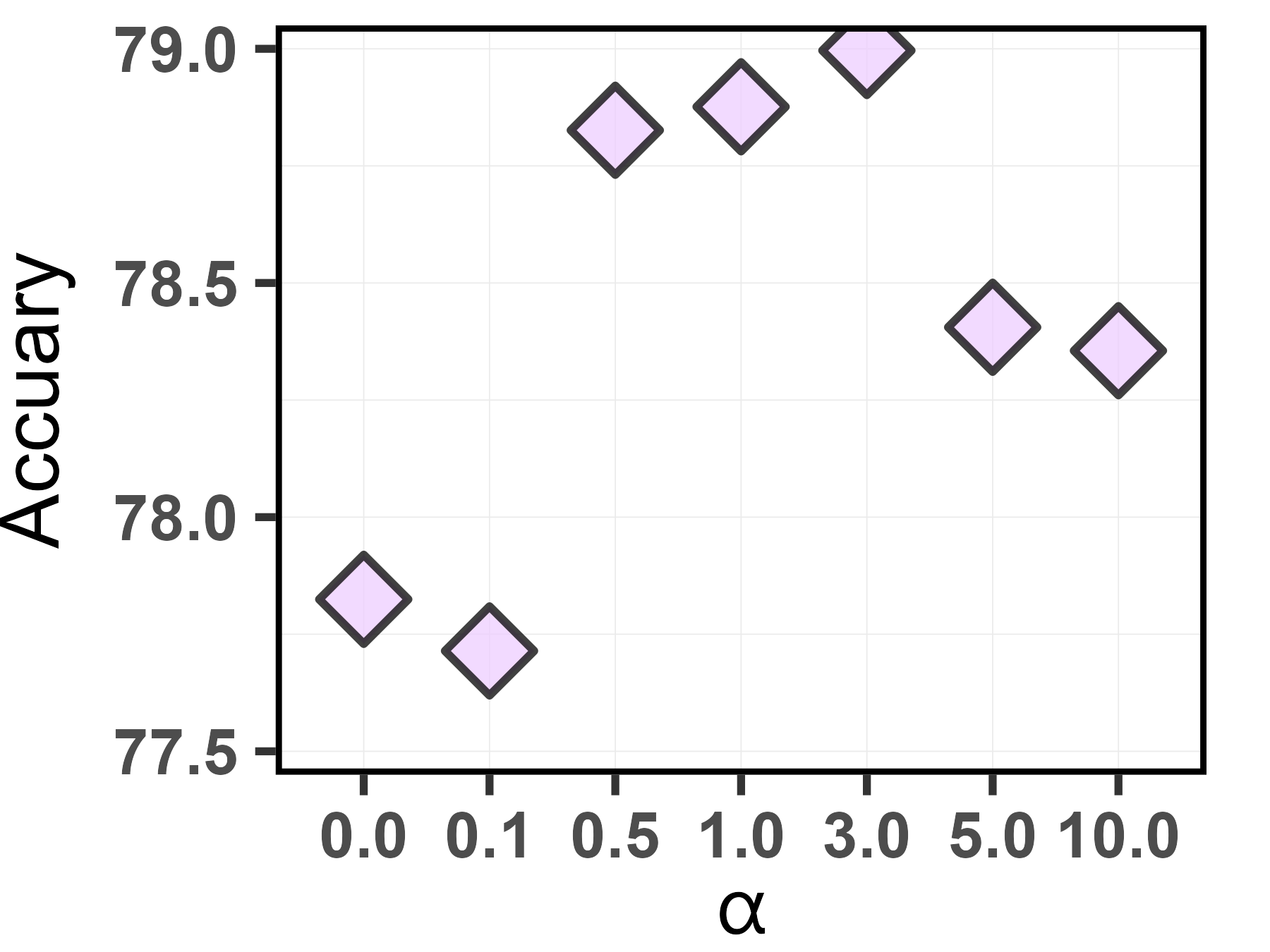}
        \label{fig:cifar100_lambda_acc}
    }
    \subfloat{
        \includegraphics[width=0.45\linewidth]{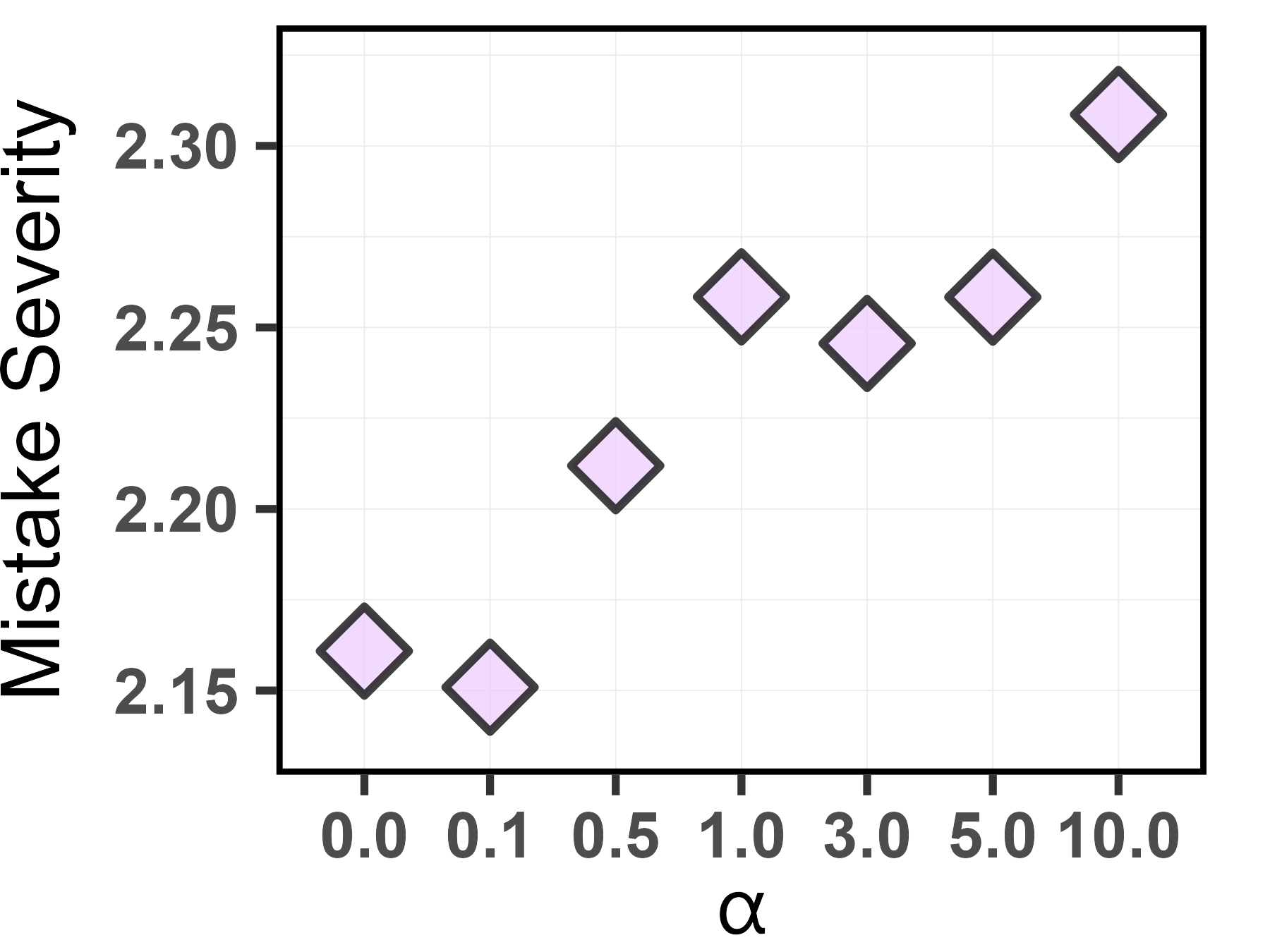}
        \label{fig:cifar100_lambda}
    }\\
    \subfloat{
        \includegraphics[width=0.45\linewidth]{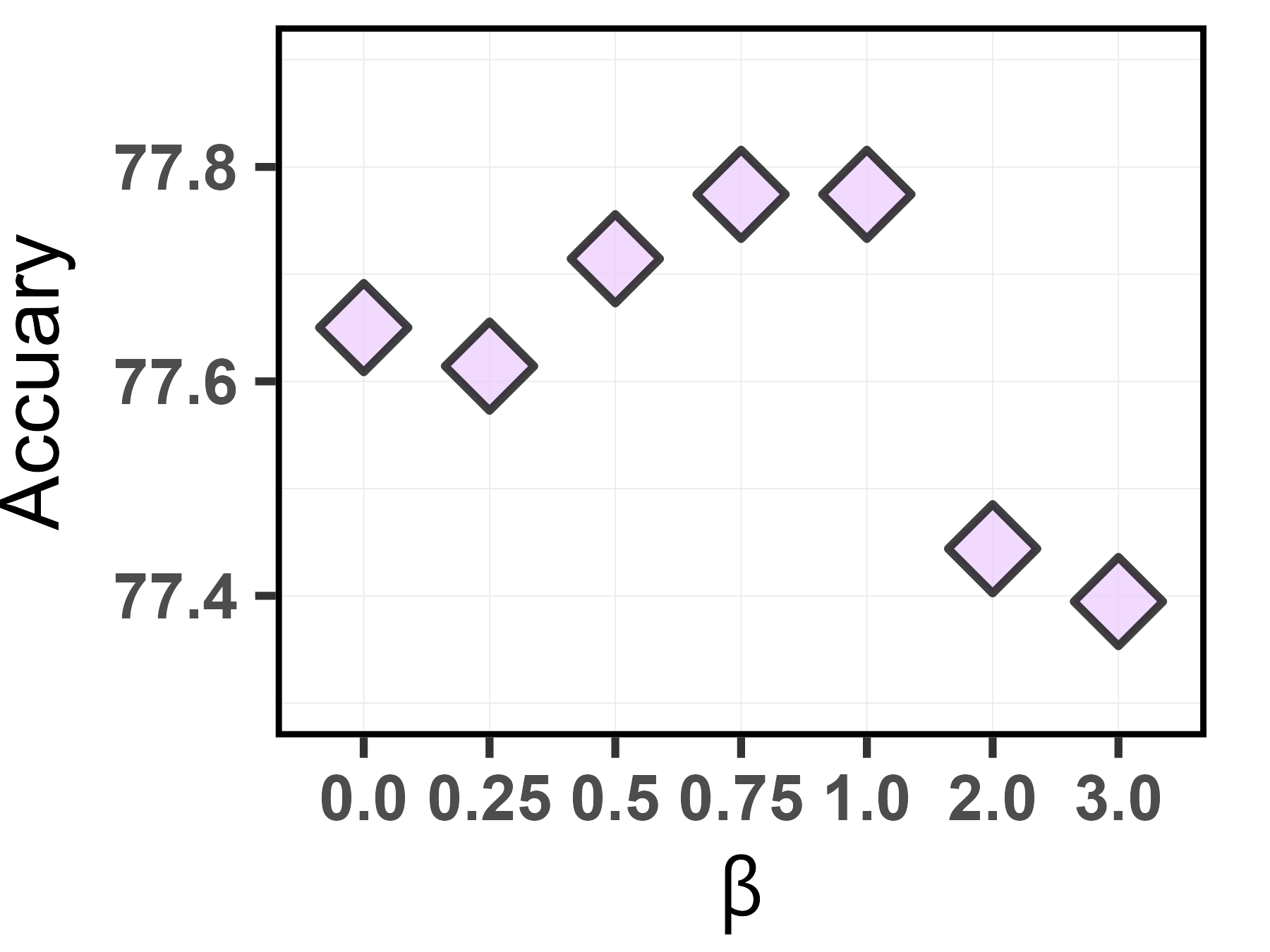}
        \label{fig:cifar100_beta_acc}
    }
    \subfloat{
        \includegraphics[width=0.45\linewidth]{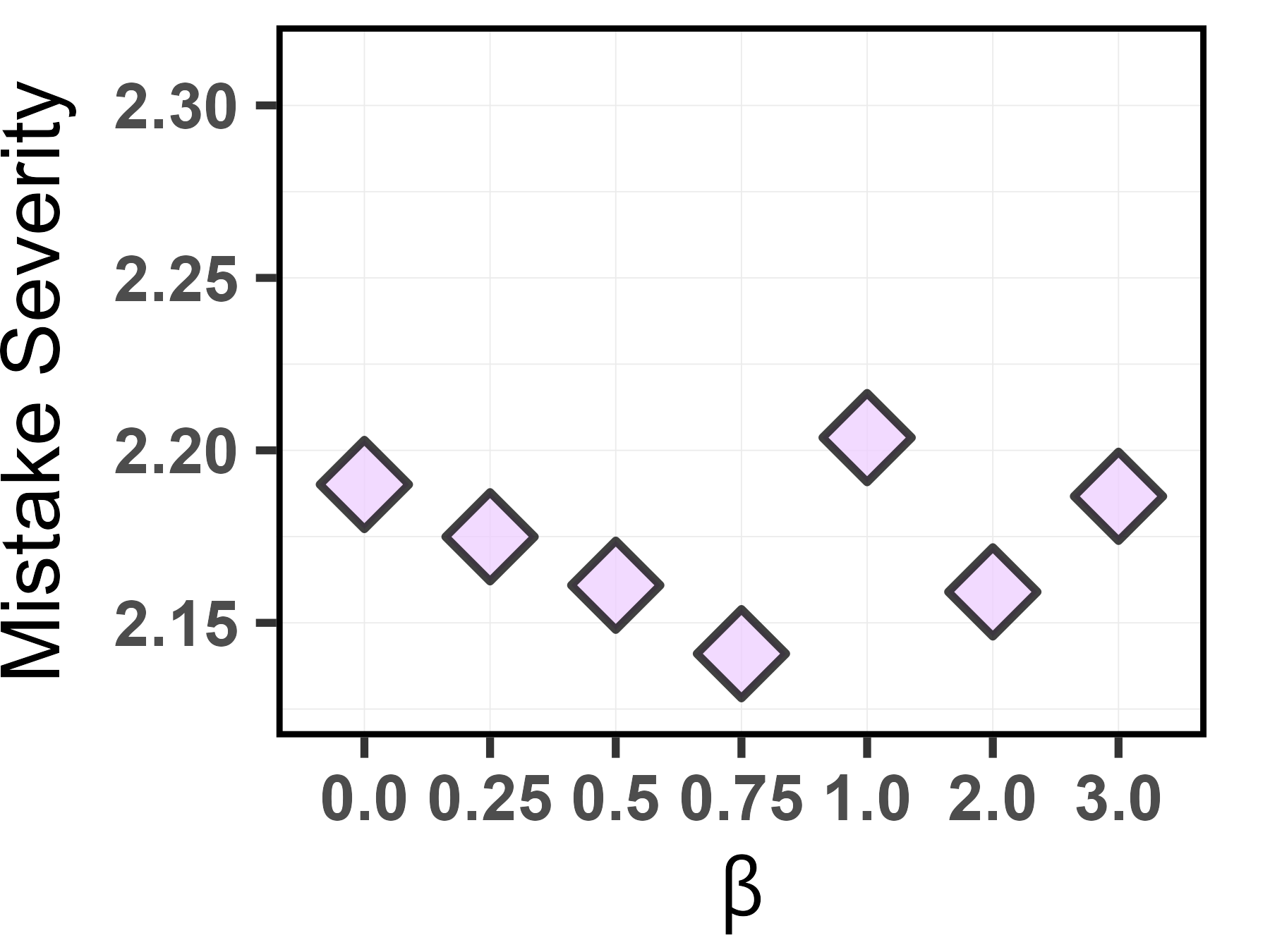}
        \label{fig:cifar100_beta}
    }
    \caption{Sensitivity analysis about $\alpha$ and $\beta$ on CIFAR-100 dataset.}
    \label{fig:cifar100_lambda_main}
\end{figure}

Fig.\ref{fig:cifar100_lambda_main}  presents the sensitivity analysis of $\beta$ on the CIFAR-100 dataset.
Like the FGVC-Aircraft dataset, $\beta$ within $0.5$ to $0.75$ can enhance fine-grained learning.

\subsubsection{Different components.}
\begin{table*}[!t]
  \centering
  \resizebox{0.95\textwidth}{!}{
        \begin{tabular}{ccc|cccc}
        \toprule
        BiLT  & AIGDL & LabelSmoothing & Mistake Severity($\downarrow$) & Hier Dist@1($\downarrow$) & Hier Dist@5($\downarrow$) & Top-1 Accuracy($\uparrow$) \\
        \midrule
                   &            &            & 2.12  & 0.44  & 2.1   & 79.35 \\
        \checkmark &            &            & 2.04  & 0.39  & 2.03  & \underline{81.07} \\
                   & \checkmark &            & 2.05  & 0.41  & \underline{1.73}  & 80.20 \\
                   &            & \checkmark & 2.10  & 0.51  & 1.81  & 75.84 \\
        \checkmark & \checkmark &            & 1.99  & 0.39  & \underline{1.73}  & 80.47 \\
        \checkmark &            & \checkmark & \underline{1.96}  & \underline{0.37}  & \textbf{1.72}  & 81.01 \\
                   & \checkmark & \checkmark & 2.05  & 0.40  & \underline{1.73}  & 80.47 \\
        \checkmark & \checkmark & \checkmark & \textbf{1.95}  & \textbf{0.36}  & \textbf{1.72}  & \textbf{81.34} \\
        \bottomrule
        \end{tabular}
    }
        \caption{Ablation study over different components of our method. The first and second best results are highlighted with \textbf{bold} text and \underline{underline}, respectively.}
  \label{tab:differentcomponents}
\end{table*}

The ablation study, which involves BiLT, AIGDL, and label smoothing, is presented in Table \ref{tab:differentcomponents} on the FGVC-Aircraft dataset.
The results show that using all three components together achieves the best performance.
When used individually, each component generally improves performance, except for Label Smoothing, which alone reduces Top-1 Accuracy.
This decline occurs because Label Smoothing, while introducing class distance information, weakens ground-truth supervision, making the model’s predictions overly conservative and reducing Top-1 Accuracy.
Among the three components, BiLT provides the most significant performance improvement, highlighting BiLT’s central role in our method.

\end{document}